\documentclass[nohyperref]{article}

\usepackage{microtype}
\usepackage{graphicx}
\usepackage{booktabs} 
\usepackage{amsmath}
\usepackage{amssymb}
\usepackage{mathtools}
\usepackage{amsthm}
\usepackage{cite}
\usepackage[utf8]{inputenc} 
\usepackage[T1]{fontenc}    
\usepackage{hyperref}       
\usepackage{url}            
\usepackage{amsfonts}       
\usepackage{nicefrac}       
\usepackage{xcolor}         
\usepackage{subcaption}
\usepackage{multirow}
\usepackage{enumitem}
\usepackage{adjustbox}
\usepackage{makecell}
\usepackage{longtable}
\usepackage{multicol,lipsum}

\usepackage{hyperref}

\usepackage[accepted]{icml2022}

\usepackage[capitalize,noabbrev]{cleveref}

\DeclarePairedDelimiter\ceil{\lceil}{\rceil}

\theoremstyle{plain}

\theoremstyle{definition}

\theoremstyle{remark}

\usepackage[textsize=tiny]{todonotes}

\icmltitlerunning{Drop Clause Tsetlin Machine}

\begin{document}

\twocolumn[
\icmltitle{Drop Clause: Enhancing Performance, Interpretability and Robustness of the Tsetlin Machine}




\begin{icmlauthorlist}
\icmlauthor{Jivitesh Sharma}{yyy}
\icmlauthor{Rohan Yadav}{yyy}
\icmlauthor{Ole-Christoffer Granmo}{yyy}
\icmlauthor{Lei Jiao}{yyy}
\end{icmlauthorlist}

\icmlaffiliation{yyy}{Center for Artificial Intelligence Research, University of Agder, Norway}

\icmlcorrespondingauthor{Jivitesh Sharma}{jivitesh.sharma@uia.no}

\icmlkeywords{Machine Learning, ICML}

\vskip 0.3in
]



\printAffiliationsAndNotice{} 

\begin{abstract}
In this article, we introduce a novel variant of the Tsetlin machine (TM) that randomly drops clauses, the key learning elements of a TM. In effect, TM with drop clause ignores a random selection of the clauses in each epoch, selected according to a predefined probability. In this way, additional stochasticity is introduced in the learning phase of TM. To explore the effects drop clause has on accuracy, training time, interpretability and robustness, we conduct extensive experiments on nine benchmark datasets in natural language processing~(NLP) (IMDb, R8, R52, MR and TREC) and image classification (MNIST, Fashion MNIST, CIFAR-10 and CIFAR-100). Our proposed model outperforms baseline machine learning algorithms by a wide margin and achieves competitive performance in comparison with recent deep learning model such as BERT and AlexNET-DFA. In brief, we observe up to $+10$\% increase in accuracy and $2\times$ to $4\times$ faster learning compared with standard TM. We further employ the Convolutional TM to document interpretable results on the CIFAR datasets, visualizing how the heatmaps produced by the TM become more interpretable with drop clause. We also evaluate how drop clause affects learning robustness by introducing corruptions and alterations in the image/language test data. Our results show that drop clause makes TM more robust towards such changes\footnote{The code is available online on: \href{https://github.com/Anonymous-2491/Drop-Clause-Interpretable-TM}{github}}.
\end{abstract}
\vspace{-0.5cm}
\section{Introduction}
\vspace{-0.1cm}
Researchers across various fields are increasingly paying attention to the interpretability of AI techniques. While interpretability previously was inherent in most machine learning approaches, state-of-the-art methods now increasingly rely on black-box deep neural network-based models. Natively, these can neither be interpreted during the learning stage nor while producing outputs. For this reason, a surge of techniques attempts to open the black box by visual explanations and gradient-based interpretability~\cite{saliency,interpretcnns,patchnet,dissection,gradcam}.\\
The Tsetlin Machine (TM) is a natively interpretable rule-based machine learning algorithm that produces logical rules~\cite{granmo2018}. Despite being logic-based, the TM is a universal function approximator, like a neural network. In brief, it employs an ensemble of Tsetlin Automata (TA) that learns propositional logic expressions from Boolean input features. Due to its Boolean representations and finite-state automata learning mechanisms, it has a minimalistic memory footprint. Propositional logic drives learning, eliminating the need for floating-point operations. More importantly, TM achieves interpretability by leveraging sparse disjunctive normal form.  Indeed, humans are particularly good at understanding flat and short logical AND-rules, reflecting human reasoning \cite{human-reasoning}.\\
However, the TM suffers from overfitting in the sense that its learning elements, i.e. clauses, are prone to capture noise and redundant patterns that are detrimental to performance. Initially, we conjectured that this could be simply because there are too many clauses. However, on decreasing the number of clauses, there was a proportionate decrease in performance. So, in order to reduce overfitting and increase generalization performance, we take inspiration from Dropout \cite{dropout} and propose a method we call \emph{Drop Clause} (DC). The drop clause method randomly \emph{drops} or switches off a set of clauses during training. This is similar to what dropout does in neural networks. However, unlike dropout, drop clause induces stochasticity in TM learning, boosting performance in terms of accuracy and learning speed. It further improves the patterns captured by the clauses, making them more interpretable and robust towards input perturbations. As a direct consequence of dropping clauses during training, the training time is reduced proportionally.
\vspace{-0.4cm}
\paragraph{Paper Contributions:} Our paper's contributions can be summarized as follows:
\vspace{-0.25cm}
\begin{enumerate}[leftmargin=0.5cm]
    \item We propose to drop clauses randomly during each TM training iteration, which introduces additional stochasticity in the learning process.
    \vspace{-0.15cm}
    \item We demonstrate that dropping clauses makes the TM capture more unique patterns, improving its generalization performance significantly.
    \vspace{-0.15cm}
    \item The improved patterns lead to competitive TM results compared to state-of-the-art models on $9$ benchmark datasets from NLP and image classification. Since TM is more akin to a standard machine learning algorithm, like a decision tree, than a deep learning model, we also compare the drop clause TM with well-known machine learning algorithms, documenting superior performance across all the datasets.
    \vspace{-0.15cm}
    \item We visualize several examples of enhanced interpretability, and measure increased robustness on image classification and natural language sentiment analysis.
\end{enumerate}
\vspace{-0.35cm}
\paragraph{Paper Organization.} The rest of the paper is laid out as follows: Section~2 introduces our proposed drop clause method. Section~3 then presents the outcomes of our NLP and image classification experiments, while Section~4 investigates the interpretability of the drop clause TM over the standard TM, both on natural language sentiment analysis and image classification. We then show how drop clause enhances TM robustness towards corrupted or noisy test data in Section~5. Section 6 contains a brief discussion of our main results and findings, with Section 7 providing the main conclusions of our research.

\section{Drop Clause for Tsetlin Machines}
As discussed in the previous section, neural networks are difficult to interpret. Therefore, applications leverage them as black boxes, with the adverse effects that entail. However, for high-stakes applications, such as healthcare, black-box approaches are not sufficient. Modern society needs reliable, unbiased, and trustworthy AI systems. Researchers have shown that neural networks are not fully mature for integration into society. For example, neural network interpretability is fragile towards adversarial examples \cite{nnfragile_vvimp}. Generating adversarial perturbations that produce visually indistinguishable images for humans leads to dramatically different neural network interpretations, without changing the label.\\
Lack of interpretability is another concern. Rudin et al. argue that neural networks are black boxes that cannot be fully explained, and offer several reasons for choosing interpretable models over attempting to explain black box ones \cite{stop}. They further raise  three algorithmic challenges that the machine learning community faces to succeed with human-level interpretability. Although neural networks are inherently inadequate for these challenges,  the TM addresses them natively:
\vspace{-0.3cm}
\begin{enumerate}[leftmargin=0.5cm]
    \item \emph{Challenge 1: Rule-based logical models}. The TM learns a rule-based model using logical operations. Learning is game-theoretic with Nash equilibria that correspond to the optimal propositional logic expressions \cite{granmo2018,39,jiao2021convergence}.
    \vspace{-0.2cm}
    \item \emph{Challenge 2: Linear models with sparse scoring systems.} The TM is a linear model that produces sparse clauses and integer weighted scores \cite{abeyrathna2021integer}.
    \vspace{-0.2cm}
    \item \emph{Challenge 3: Domain-specific interpretable AI.} The TM is inherently interpretable, used for producing interpretable models across several domains \cite{abeyrathna2020intrusion,berge2019text,lei2021kws,rohan2}.
\end{enumerate}
\vspace{-0.25cm}
In this section, we first briefly present the basic TM as well as the convolutional version. Thereafter, we introduce the details of our drop clause technique, which enhances the stochasticity of TM learning. Our goal is to reduce overfitting behaviour and considerably increase performance. We also show enhanced interpretability on CIFAR-10 at the pixel-level and on MR dataset at the word-level. And, as the TM is based on Boolean data and propositional logic, it has very low memory footprint and computational complexity \emph{(zero FLOPs)}. This makes the usage of TM advantageous for on-device and federated learning on mobile devices and devices with limited compute.

\subsection{TM and Convolutional TM}
A TM in its simplest form takes a feature vector $\mathbf{x} = [x_1, x_2, \ldots, x_o] \in \{0,1\}^o$ of $o$ Boolean values as input, producing a Boolean output $\hat{y} \in \{0,1\}$ \cite{granmo2018}. Patterns are expressed as conjunctive clauses~(AND-rules), built from literals $L = \{l_1, l_2, \ldots, l_{2o}\} = \{x_1, x_2, \ldots, x_{o}, \lnot x_1, \lnot x_2, \ldots, \lnot x_o\}$:
\vspace{-0.2cm}
\begin{multline}
    \hat{y} = 0 \le \sum_{j=1,3,\ldots}^{n-1} \bigwedge_{k=1}^{2o} \left[g(a_k^j) \Rightarrow l_k\right] \\ - \sum_{j=2,4,\ldots}^{n} \bigwedge_{k=1}^{2o} \left[g(a_k^j) \Rightarrow l_k\right]. \label{eqn:1}
\end{multline}
In Equation \ref{eqn:1}, $a^j_k$ is the TA state that controls inclusion of literal $l_k$ in clause $j$ and $g(\cdot)$ maps the TA state to action $0$ or $1$. The imply operator, in turn, implements the action in the clause. Even-indexed/odd-indexed clauses vote for output $\hat{y}=0$/$\hat{y}=1$. A detailed explanation can be found in the Appendix~A.1.\\
The Convolutional TM (CTM) is as an interpretable alternative to CNNs \cite{granmo2019convtsetlin}. Whereas the TM categorizes an image by employing each clause once to the whole image, the CTM uses each clause as a convolution filter. That is, a clause is evaluated multiple times, once per image patch taking part in the convolution. The output of a convolution clause is obtained simply by ORing the outcome of evaluating the clause on each patch:
\begin{multline}
    \hat{y} = 0 \le \sum_{j=1,3,\ldots}^{n-1}  \bigvee_{b=1}^B \left[\bigwedge_{k=1}^{2o} \left[g(a_k^j) \Rightarrow l_k^b\right] \right] \\ - \sum_{j=2,4,\ldots}^{n} \bigvee_{b=1}^B \left[ \bigwedge_{k=1}^{2o} \left[g(a_k^j) \Rightarrow l_k^b\right]\right]. \label{eqn:prediction_convolution}
\end{multline}
Here, $b$ refers to one out of $B$ available image patches \cite{granmo2019convtsetlin}. See Appendix A.2 for further details.

\subsection{Drop Clause}
Although there might be an enormous amount of patterns in real-life data, with a sufficient number of clauses the TM can identify the most important ones. However, patterns might differ between training and testing data, resulting in overfitting. We therefore propose a novel regularization method for the TM called drop clause. This technique is inspired by the dropout method for neural networks. In drop clause, clauses are removed with a probability $p$ in each training epoch:
\vspace{-0.2cm}
\begin{multline}
    \hat{y} = 0 \le \sum_{j=1,3,\ldots}^{n-1} \pi_j \bigwedge_{k=1}^{2o} \left[g(a_k^j) \Rightarrow l_k\right] \\ - \sum_{j=2,4,\ldots}^{n} \pi_j \bigwedge_{k=1}^{2o} \left[g(a_k^j) \Rightarrow l_k\right]. \label{eqn:prediction}
\end{multline}
Above, $\pi_j \in \{0,1\}$ for clause $j$ is zero with probability $p$ for each complete epoch. The purpose is to reduce the chance of learning redundant patterns. Accordingly, the vanilla TM and its drop clause variant are equivalent if $p=0$.\\
Drop clause in TM works in a similar way as dropout in neural networks. A basic TM seeks to minimize prediction error on the training data. Achieving this, there may still be unused patterns available in the training data. These patterns can potentially be useful when facing new data, such as test data. By randomly dropping clauses for complete epochs, we mobilize other clauses to take over the role of the dropped clauses. Drop clause can be thought of as having a pool of clauses and selecting a set of clauses with $1-p$ probability for training every epoch. However, due to this stochastic learning, the mobilized clauses may solve the task in a different way.  Hence, robustness increases overall when the complete set of clauses are turned on again. The resulting effects are evaluated experimentally in the next section.\\
The stochasticity induced by dropping clauses in the TM learning process is in some senses similar to the stochasticity induced by stochastic gradient descent (SGD) \cite{sgd}. In SGD, a mini-batch of training samples is selected  randomly from the training data. While, in drop clause, various sets of clauses are selected for training with probability $1-p$. The major difference is that, SGD is stochastic in terms of data sampling during the learning stage, whereas drop clause is stochastic in terms of sampling the elements of the model structure (clauses are the elements that form the TM model). Another difference is that the stochasticity of drop clause can be adjusted with the drop clause probability hyper-parameter $p$. Note that even with stochastically and randomly reduced number of clauses, that change ever training epoch, the TM still retains its convergence properties. Please see Appendix A.3 for details.

\section{Experiments and Results}
We here investigate the effects drop clause has on the performance of TMs. To assess the generality of our approach, we test drop clause both on NLP and image classification. For NLP, we use TM with weighted clauses, and for image classification, we use the CTM. For evaluation, we compare our approach with comparable state-of-the-art deep learning models. Since TM is a rule-based machine learning method, we also cover more standard machine learning algorithms. The drop clause implementation is written in PyCUDA for parallel GPU computation. We train our models on 16 NVIDIA Tesla V100 TensorCore GPUs for the image classification task and on one NVIDIA RTX 3070 8GB GPU for the NLP tasks. 

\subsection{Natural Language Processing}
We explore the performance of drop clause on NLP tasks first. The first experiments assess how varying the drop clause probability affects classification accuracy and then compare the best found configuration with similar machine learning algorithms and state-of-the-art techniques. To this end, we use $5$ popular standard datasets --- IMDb, R8, R52, MR and TREC, summarized below: 
\vspace{-0.35cm}
\begin{itemize}[leftmargin=0.5cm]
    \item \emph{\textbf{IMDb}} consists of $50,000$ movie reviews for binary sentiment analysis, out of which $25,000$ are used for training and $25,000$ for testing. Here, we use $10,000$ clauses and set the TM hyperparameters target  $T$ to $8,000$ and specificity $s$ to $2.0$ (see Appendix~A.1 for details).
    \vspace{-0.25cm}
    \item \emph{\textbf{Reuters-21578}} contains the text categorization datasets \emph{\textbf{R8}} and \emph{\textbf{R52}}. R8 is divided into $8$ categories, with $5,485$ training and $2,189$ testing samples, whereas R52 consists of $52$ categories, divided into $6,532$ training and $2,568$ testing samples. We here employ $3,000$ clauses with $T=2,000$ and $s=7.0$.
    \vspace{-0.25cm}
    \item \emph{\textbf{MR}} is another movie review dataset for binary sentiment classification, consisting of $10,662$ samples. We use the train-test split as in \cite{mr-split} and use $5,000$ clauses with $T=4,000$ and $s=6.0$.
    \vspace{-0.25cm}
    \item \emph{\textbf{TREC}} is a question classification dataset that encompasses $6$ categories. There are a total of $6,000$ samples, $5,500$ for training and $500$ for testing. Here, we use $5,000$ clauses with $T=4,000$ and $s=2.0$.
\end{itemize}
\vspace{-0.35cm}
Our first step is to compare the difference in performance with respect to change in drop clause probability, shown in Table~\ref{tab:pnlp}. The selected drop clause probabilities are $p \in \left\{0.1, 0.25, 0.5, 0.75\right\}$. For IMDb, R8, R52 and MR, the best performance is achieved with $p=0.75$, which is equivalent of dropping $75\%$ of the total clauses per class, per epoch. As for TREC, $p=0.5$ works best. As seen, drop clause has significant effect on accuracy, with average accuracy going up by $4.07\%$ for IMDb, $2.78\%$ for R8, $3.61\%$ for R52, $3.53\%$ for MR and $2.45\%$ for TREC. Additionally, we observe a substantial reduction in training time proportional to the drop clause ratio. Inference times on these datasets are less than $2.3ms$ per sample on average on an NVIDIA RTX 3070 8GB GPU. \\
\begin{table}[t]
\caption{Effect of Drop Clause on NLP Datasets.}
\label{tab:pnlp}
\begin{center}
\begin{small}
\begin{sc}
\adjustbox{max width=\textwidth}{%
\begin{tabular}{lccccc}
\toprule
TM & $p=0$ & $p=.1$ & $p=.25$ & $p=.5$ & $p=.75$ \\
    \midrule
    IMDB &   87.2 & 88.3  & 89.6  & 90.4  & \textbf{91.27} \\
    R8 & 96.16  &  97.6 &  98.1  &  98.5  &  \textbf{98.94} \\
    R52 &  89.14 &  89.5  & 90.8   &   91.5 &  \textbf{92.75} \\
    MR &  75.14  &  77.25  & 77.9   &  78.2  &  \textbf{78.67} \\
    TREC &  88.05  &  89.8  &  90.1  &  \textbf{90.5}  &  89.9 \\
\bottomrule
\end{tabular}}
\end{sc}
\end{small}
\end{center}
\vskip -0.25in
\end{table}
\noindent The TM is a rule-based machine learning algorithm, so we also compare our proposed TM model with a few traditional machine learning techniques. Table~2 displays the results of our comparison and shows TM's superior performance among its class of algorithms\footnote{Note that the machine learning methods used here for comparison have default parameters from sklearn.}. We compare our proposed TM with drop clause against Support Vector Machine (SVM), Random Forests (RF), K-Nearest Neighbours (K-NN) and XGBoost (XGB).  From Table 2, it is clear that TM with drop clause outperforms other machine learning algorithms by a wide margin.\\
We also compare our model with deep learning methods, some of them representing state-of-the-art. Tables \ref{tab:sotanlp} and \ref{tab:sotanlp2} show the results of our comparisons. Enhanced with drop clause, the accuracy of TM is comparable not only to CNNs and LSTMs but also to computationally complex and parametrically large state-of-the-art models like BERT. On all datasets in Table \ref{tab:sotanlp}, inclusion of drop clause propels the performance of the TM to outperform CNN \cite{cnnn} and bidirectional LSTM with pretrained word embeddings. In fact, on R8 and R52, drop clause TM is able to achieve better accuracy than BERT \cite{bert} and comes close to its performance on MR. Also, comparing our model with the state-of-the-art graph convolutional neural network, $S^2GC$ \cite{s2gc}, drop clause TM achieves better performance on R8 and MR, and comparable performance on R52. Similar results are obtained on the TREC dataset shown in Table \ref{tab:sotanlp2}. Drop clause TM outperforms a baseline LSTM model as well as a vanilla Transformer (TF) and Transformer with feature projection (FP) \cite{fp}.\\
Even though BERT achieves better performance on TREC dataset, TM has major advantages over these models concerning computational complexity and interpretability, which are further enhanced by drop clause (See Section 5). While it is disputed whether attention is explainable \cite{notxai}, the approach is significantly more complex than our proposed model and achieves comparable performance. Note that our model only relies on simple bag-of-words tokens in the target datasets, without considering any additional pretrained world knowledge, like word2vec, Glove or BERT features. Yet, it achieves competitive performance compared to deep learning and some state-of-the-art models. 

\begin{table}[t]
\caption{Comparison with Machine Learning methods}
\label{tab:cnlp}
\begin{center}
\begin{small}
\begin{sc}
\adjustbox{width=.47\textwidth}{%
\begin{tabular}{lcccccc}
\toprule
 & TM & \multicolumn{1}{c}{TM} & SVM & RF & K-NN & XGB\\
  &  & (DC) &  &  &  &  \\  \midrule
    IMDB &   87.2 & 91.27  & 83.2 & 78.5 & 72.5 & 84.4 \\
    R8 & 96.16 & 98.94 & 84.7 & 82.4 & 77.3 & 85.1 \\
    R52 & 89.14 & 92.75 & 71.0 & 72.5 & 58.2 & 75.4 \\
    MR &  75.14 & 78.67 & 52.2 & 51.7 & 39.1 & 56.3 \\
    TREC &  88.05 & 90.5 & 67.5  &  65.4  & 56.1  &  77.6 \\
\bottomrule
\end{tabular}}
\end{sc}
\end{small}
\end{center}
\vskip -0.1in
\end{table}

\begin{table}[t]
\caption{Comparison on IMDb, R8, R52 and MR}
\label{tab:sotanlp}
\begin{center}
\begin{small}
\begin{sc}
\begin{tabular}{lccccc}
\toprule
 & \multicolumn{1}{c}{TM} & CNN & BiLSTM & BERT & S$^2$GC\\
 & (DC) &  &  &  &  \\
    \midrule
    IMDB & 91.27 & 87.9 &88.9 & 95.4 & -\\
    R8  & 98.94 & 95.71 &96.31 & 96.02 & 97.4 \\
    R52  & 92.75 & 87.59 &90.54 & 89.66 & 94.5 \\
    MR & 78.67 & 77.75 &77.68 & 79.24 & 76.7 \\
\bottomrule
\end{tabular}
\end{sc}
\end{small}
\end{center}
\vskip -0.1in
\end{table}

\begin{table}[t]
\caption{Comparison on TREC-6}
\label{tab:sotanlp2}
\begin{center}
\begin{small}
\begin{sc}
\begin{tabular}{lcccccc}
\toprule
 & \multicolumn{1}{c}{TM} & LSTM & TF & FP+TF & BERT\\
 & (DC) &  &  &  &  \\
    \midrule
    TREC-6 & 90.5 & 87.19 & 87.33 & 89.5 & 95.6\\
\bottomrule
\end{tabular}
\end{sc}
\end{small}
\end{center}
\vskip -0.1in
\end{table}

\subsection{Image Classification}
We now turn to image classification, again focusing on how drop clause affects performance of TM. To this end, we evaluate drop clause on four benchmark image classification datasets: MNIST, Fashion-MNIST, CIFAR-10 and CIFAR-100. We binarize the datasets using an adaptive Gaussian thresholding procedure as proposed in \cite{granmo2019convtsetlin}. This binarization results in images with merely $1$ bit per pixel per channel, considerably reducing the memory overhead. For the task of image classification, we employ the Convolutional TM (CTM).
\vspace{-0.35cm}
\par \begin{itemize}[leftmargin=0.5cm]
    \item \emph{\textbf{MNIST}} encompasses $70,000$ $28\times28$ gray-scale   images of hand written single digits, $60,000$ for training and $10,000$ for testing. Here, we use $8,000$ clauses with $T=6,400$ and $s=5.0$.
\vspace{-0.25cm}
    \item \emph{\textbf{Fashion-MNIST}} contains $28\times28$ gray-scale images from the Zalando catalogue. For this dataset, we use the same parameters as the MNIST dataset.
    \vspace{-0.25cm}
    \item \emph{\textbf{CIFAR-10 and CIFAR-100}} consist of $50,000$ $32\times32$ color images of objects divided into $10$ categories for CIFAR-10 and $100$ for CIFAR-100. Another $10,000$ images are provided for testing. Here, we make use of $60,000$ clauses with $T=48,000$ and $s=10.0$.
\end{itemize}
\vspace{-0.25cm}
Again, we explore the effects of four drop clause probability settings, $p:\left\{0.1, 0.25, 0.5, 0.75\right\}$, shown in Table \ref{tab:p}. The best performance is achieved with $p=0.25$ for the MNIST and Fashion-MNIST, while $p=0.5$ gives the best result for the CIFAR datasets, which is equivalent to dropping a quarter (or half) of the clauses per training iteration. Notice the considerable performance increase, especially on the CIFAR datasets, with drop clause of $p=0.5$. Peak accuracy on MNIST is $99.45\%$ $(\pm0.25\%)$, Fashion-MNIST is $92.5\%$ $(\pm0.25\%)$. CIFAR-10 and CIFAR-100 peak at $75.1\%$ $(\pm0.4\%)$ and $45.2\%$ $(\pm0.2\%)$ respectively, averaged over $100$ runs\footnote{We also tested our model on Kuzushiji-MNIST. Our best model achieved $97.6\%$ $(\pm0.2\%)$, averaged over 100 runs, which is close to ResNet-18's $97.82\%$. We here used the same hyperparameters as for MNIST.}. Apart from the accuracy gain, drop clause with $p=0.5$ reduces training time by approximately $50\%$, for the CIFAR datasets. For instance, training time per epoch drops from $61.83 s$ to $32.58 s$ for CIFAR-10. The inference time for the drop clause CTM on CIFAR-10 is $1.5 ms$ per image on one NVIDIA Tesla V100 GPU, leveraging the computation benefits of binary operations for TM inference, and no floating point operations.
\begin{table}[t]
\caption{Effect of Drop Clause on Image Classification.}
\label{tab:p}
\begin{center}
\begin{small}
\begin{sc}
\adjustbox{width=0.49\textwidth}{%
\begin{tabular}{lccccc}
\toprule
 CTM & $p=0$ & $p=.1$ & $p=.25$ & $p=.5$ & $p=.75$ \\
    \midrule
    MNIST &   99.3 & 99.3  & \textbf{99.45}  & 99.35  & 98.2 \\
    F-MNIST &    91.5 &  91.75 &  \textbf{92.5}  &  92.25  &  91.25 \\
    CIFAR-10 &  69.3  & 70.5  &  73.2 & \textbf{75.1}  & 72.6 \\
    CIFAR-100 & 35.5   & 39.5 & 42.6  & \textbf{45.2} &   40.8 \\
\bottomrule
\end{tabular}}
\end{sc}
\end{small}
\end{center}
\vskip -0.1in
\end{table}
Table \ref{tab:c} shows the comparison of the drop clause CTM with other popular machine learning techniques on the image classification problem. As seen, CTM outperforms the traditional machine learning techniques by a wide margin, further widened by the introduction of drop clause\footnote{Note that the accuracy on CIFAR-100 is not displayed in Table 6 for other machine learning algorithms as the accuracy is miniscule}.\\
Table \ref{tab:sota} contains a comparison between the drop clause CTM with some related and state-of-the-art techniques. Binary neural networks can be represented exactly as a propositional logic expression \cite{bnn2plogic}. As such, they are particularly comparable to CTM and we therefore also include results for EEV-BNN \cite{bnncompare}. EEV-BNN is a binary neural network that has been verified with Boolean satisfiability. Employing EEV-BNN, we use the MNIST-MLP and Conv-Large BNN architectures from \cite{bnncompare} for comparisons. The drop clause CTM significantly outperforms EEV-BNN on MNIST and CIFAR-10, as shown in Table \ref{tab:sota}. We also compare our method with a transformer based model of reduced complexity to make it more comparable to our model. The Vision Nystromformer~(ViN) \cite{vin} is a vision transformer~(ViT) based model that reduces the quadratic computational complexity of ViT by using the Nystrom methods for approximating self-attention. It achieves $65.06\%$ accuracy on CIFAR-10 whereas the Hybrid-ViN, which incorporates rotary positional embedding, obtains slightly better accuracy as shown in Table \ref{tab:sota}, comparable to our model.\\
We also contrast our model against AlexNet with Direct Feedback Alignment~(DFA) \cite{dfa} for parallel backpropagation training, which makes it comparable to the fast training of TM. The results in Table \ref{tab:sota} show that AlexNet-DFA achieves similar performance on MNIST and Fashion-MNIST, whereas it is outperformed by drop clause CTM on CIFAR-10. However, there is a considerable difference on the CIFAR-100 dataset due to larger label space. Finally, as drop clause is inspired from dropout, we compare our model with the Maxout network \cite{maxout}, which was a natural companion to dropout. Maxout achieves significantly better performance on the CIFAR datasets as shown in Table \ref{tab:sota}. Note that on color images, binarization can lead to loss of important information especially as we use 1-bit per channel, compared to 8-bits per channel for lossless color information. Also, as previously stated, TM possesses computational advantages over these methods along with interpretability.\\
\begin{table}[t]
\caption{Comparison on Image Classification with Machine Learning methods}
\label{tab:c}
\begin{center}
\begin{small}
\begin{sc}
\adjustbox{width=0.48\textwidth}{%
\begin{tabular}{lccccccc}
\toprule
 & CTM & \multicolumn{1}{c}{CTM} & SVM & RF & K-NN & XGB \\
    &   &  (DC) &  &  &  &  \\
    \midrule
    MNIST &   99.3 & 99.45  & 93.4  & 93.8  &  95.3 &  96.2 \\
    F-MNIST & 91.5 & 92.5 & 84.6 & 87.5  &  85.4 &  88.4 \\
    CIFAR-10 &  69.3  & 75.1  &  37.5  & 48.7  &  33.9 & 47.8 \\
    CIFAR-100 & 35.5 & 45.2 & - & - & - & - \\
\bottomrule
\end{tabular}}
\end{sc}
\end{small}
\end{center}
\vskip -0.1in
\end{table}
\begin{table}[h]
\caption{Comparison on Image Classification with different SOTA models}
\label{tab:sota}
\begin{center}
\begin{small}
\begin{sc}
\adjustbox{width=0.48\textwidth}{%
\begin{tabular}{lccccc}
\toprule
  & \multicolumn{1}{c}{CTM} & EEV & Hybrid & AlexNet & Maxout \\
  &  (DC) & -BNN & -ViN & -DFA &  \\
    \midrule
    MNIST & 99.45  & 98.25  & 96.4  & 98.2  &  99.06 \\
    F-MNIST & 92.5 & - & - & 91.66 & - \\
    CIFAR-10 & 75.1  &  63.45 & 75.26  & 64.7  &  86.8 \\
    CIFAR-100 & 45.2 & - & - & 52.62 & 59.52\\
\bottomrule
\end{tabular}}
\end{sc}
\end{small}
\end{center}
\vskip -0.1in
\end{table}
Another point to note is that, originally, dropout can increase the performance of basic CNNs by about $3\%$ and $6\%$ on CIFAR-10 and CIFAR-100 respectively \cite{dropout}. The drop clause method shows promising results by improving the performance of vanilla CTM by about $6\%$ and $10\%$ on CIFAR-10 and CIFAR-100, respectively.

\section{Enhanced Interpretability}
In this section, we show how drop clause is able to enhance the interpretable maps produced by TM and CTM for NLP and image classification tasks. We show interpretable results on the MR dataset at the word-level and on the CIFAR-10 dataset at the pixel-level. Enhancement in interpretability is in terms of better word representations for NLP tasks and more accurate pixel representations for objects in images.

\subsection{Natural Language Sentiment Analysis}
\begin{figure*}[h]
\begin{center}
\begin{subfigure}{0.45\linewidth}
  \includegraphics[width=\linewidth, height=6cm]{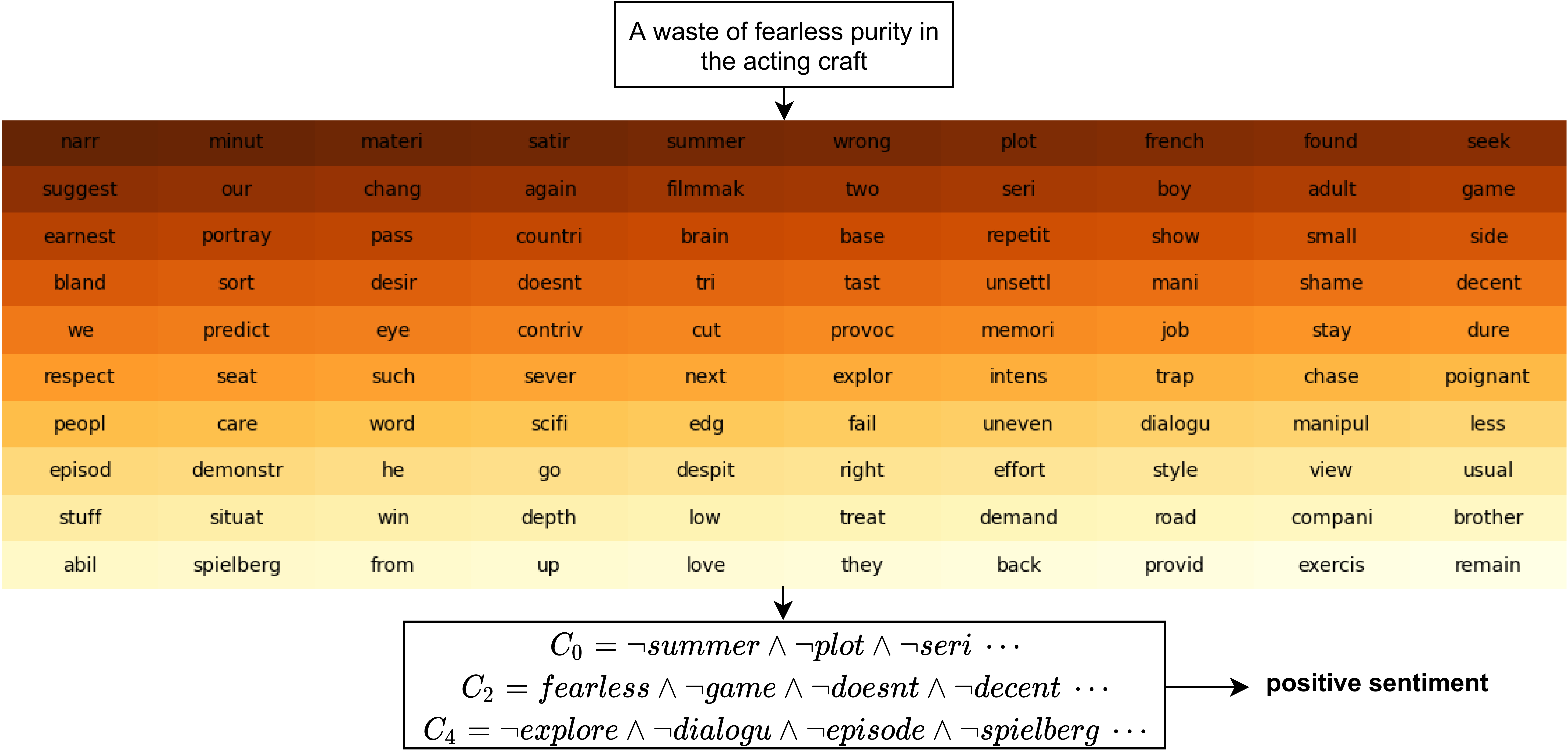}
  \caption{Without drop clause}
  \label{figndc}
\end{subfigure}
\hspace{1cm}
\begin{subfigure}{0.45\linewidth}
  \includegraphics[width=\linewidth, height=6cm]{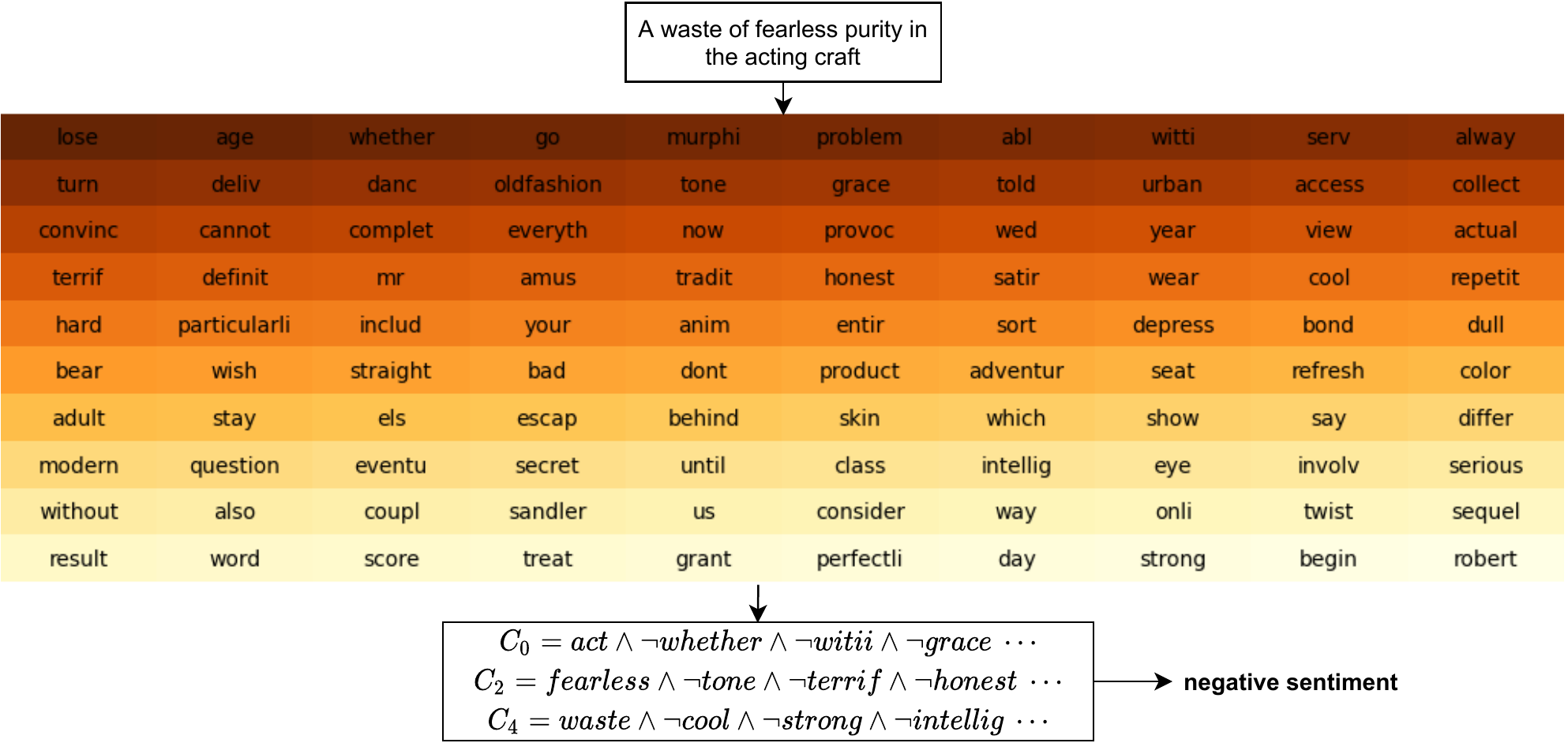}
  \caption{With drop clause $(p=0.75)$}
  \label{figdc}
  \end{subfigure}
  \caption{NLP Interpretability of a sample: (a) without and (b) with drop clause.} 
  \end{center}
\end{figure*}

We investigate local model interpretability in NLP by using a randomly selected test example from MR: "A waste of fearless purity in the acting craft" (additional examples can be found in Appendix C). The example was selected among the ones that where correctly classified with drop clause and incorrectly classified without. The purpose is to contrast how drop clause affects interpretability. Note that for NLP, the majority of the features appear in negated form. In order to have  word-level understanding of the model, we use a frequency-based interpretation, i.e., we highlight the features based on how frequently they appear in clauses, as explained in \cite{rohan2}. An arguable unique property of TMs is that the patterns they produce are both descriptive (frequent) and discriminative. In other words, each clause captures a full description of the target concept, not merely the discrimination boundary.\\
Figure \ref{figndc} highlights the top 100 most frequent features in negated form (bright color means high frequency),  present in the clauses triggered by the given sample. Three example clauses $C_0$, $C_2$, and $C_4$ are shown below for the color-coded features. In this case, the prediction is wrong. More importantly, the literals in the conjunctions do not make sense.\\
The corresponding result for drop clause is depicted in Figure \ref{figdc}. Here, the $100$ most frequent features in the negated form seem intuitive for predicting negative sentiment. Features such as "NOT witti (witty)", "NOT grace (graceful)", "NOT terrif (terrific)", "NOT honest", "NOT cool", or "NOT intellig (intelligent)" generally mean absence of positive sentiment, which in this case makes the model draw the correct conclusion (negative sentiment). Some of the specific patterns that are responsible for predicting the correct output are $C_0, C_2,$ and $C_4$, as shown in the figure. Although randomly chosen, this example is representative for how TM with drop clause is able to capture a larger variety of correct patterns than what the vanilla TM is capable of (further exemplified in Appendix C).

\subsection{Image Classification}
\begin{figure}[h]
\begin{center}
\begin{subfigure}{0.33\linewidth}
  \centering
  \includegraphics[width=1.2\textwidth]{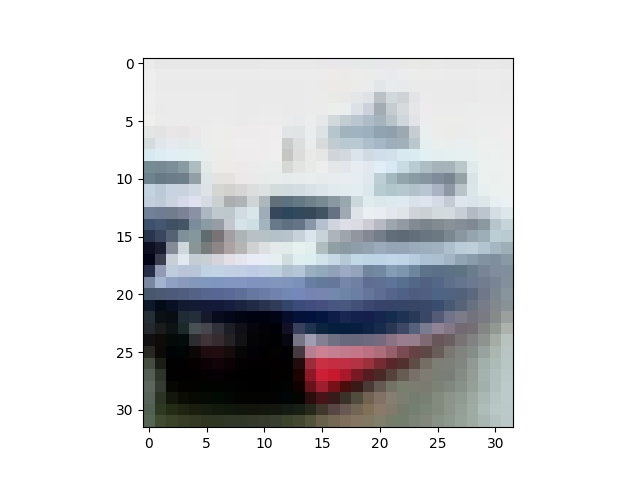}
  \caption{Image}
  \label{fig:sfig1}
\end{subfigure}%
\begin{subfigure}{0.33\linewidth}
  \centering
  \includegraphics[width=1.2\textwidth]{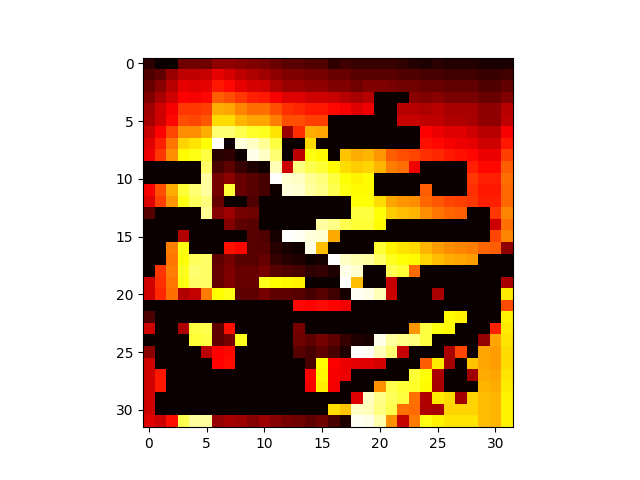}
  \caption{TM $(p=0)$}
  \label{fig:sfig2}
\end{subfigure}%
\begin{subfigure}{0.33\linewidth}
  \centering
  \includegraphics[width=1.2\textwidth]{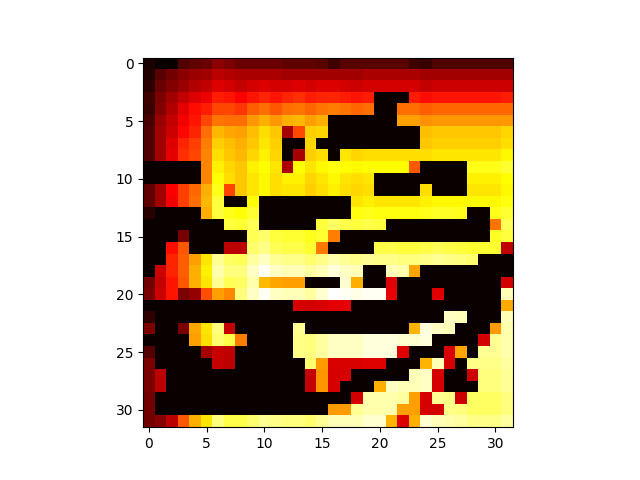}
  \caption{TM $(p=0.5)$}
  \label{fig:sfig5}
\end{subfigure}
\caption{CIFAR-10 Interpretability: (b) without and (c) with DC}
\label{fig:cifar_interpret_dc}
\end{center}
\end{figure}

\begin{figure*}[h]
  \begin{center}
  \includegraphics[width=0.7\textwidth, height=7cm]{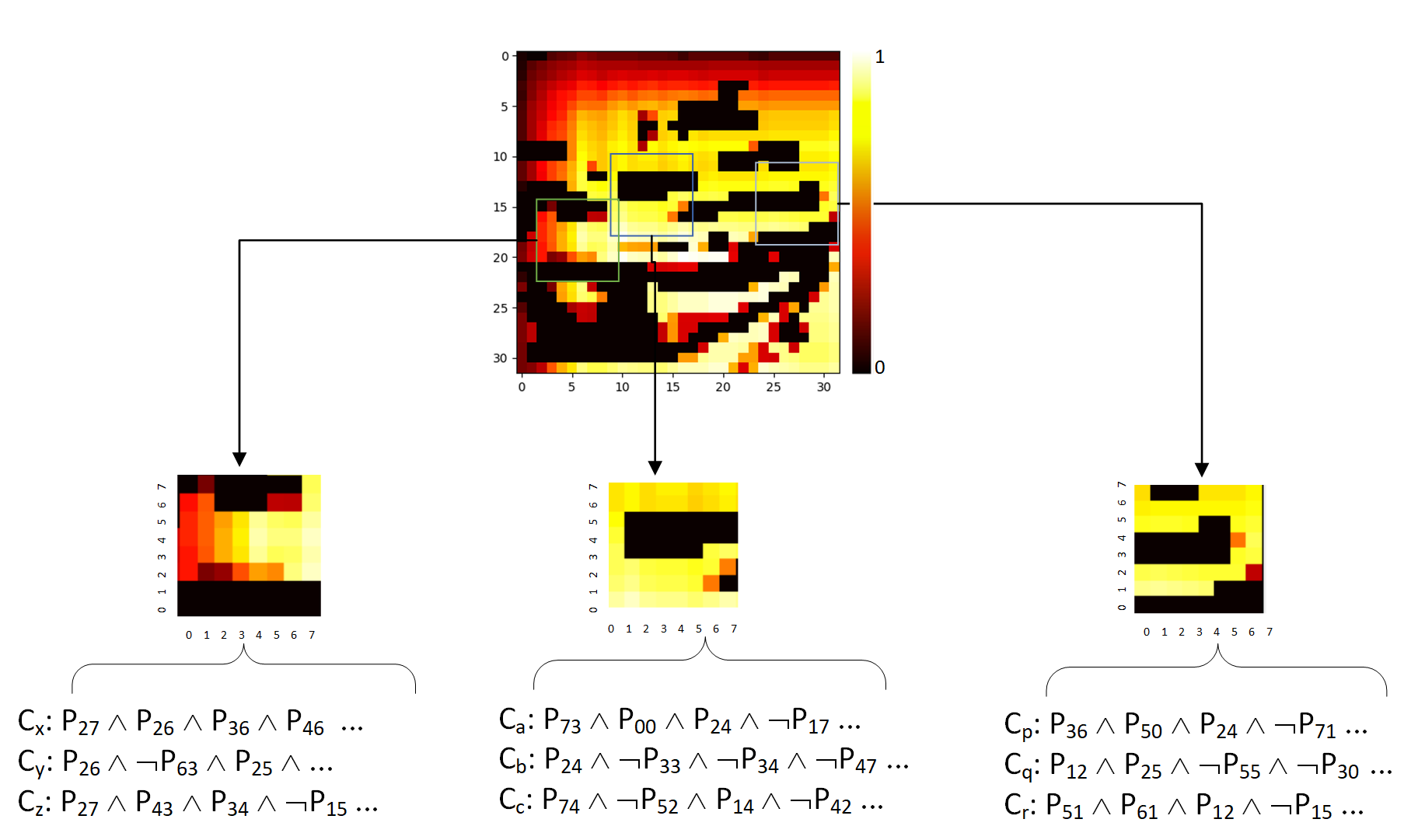}
  \caption{CIFAR-10 Pixel-level Interpretability - \emph{Patches of $8\times8$ are taken at a time. The three clauses shown per patch are some of the top 100 highest weighted clauses for the \emph{ship} class that have maximum activations in that patch. The pixels shown in the clauses are activated in that patch. Note that, the clauses are longer than shown here.}} 
  \label{fig:cifar_pixel_interpret}
  \end{center}
\end{figure*}

In image classification, TM clauses form self-contained patterns by joining pixels into multi-pixel structures. That is, the image pixels are inputted directly to the clauses, that are propositional AND-rules. So, Boolean propositional expressions (clauses) capture patterns in an image which contain pixels as literals in the clauses, that are comparatively easy for humans to comprehend \cite{valiant1984learnable}.\\
The CTM forms clauses from the pixels of the square image patches obtained in the convolution. Accordingly, we extract the top $k$ weighted clauses per class in patch form. For visualization, we represent the non-negated pixels of a clause as $1.0$, negated pixels as $-1.0$, and excluded pixels as $0.0$. In additional to the image content, each clause also encodes positions in the image where it is valid. If a certain position is invalid, the clause literal corresponding to the pixel values is treated as $0.0$ to indicate no activation at that location. The resulting clause masks are applied to the image and their activation maps are added up to produce a heatmap.\\
In Figure \ref{fig:cifar_interpret_dc}, we compare the heatmaps produced by the vanilla CTM and the CTM with drop clause (more such comparisons between image heatmaps can be found in Appendix C). As can be seen from the figure, drop clause is able to capture the object with slightly more precision. We believe this is because the remaining clauses are forced to substitute the dropped clauses, learning to perform their tasks. Due to the stochastic nature of learning, they will, however, learn to perform the tasks differently than the dropped clauses. As a result, drop clause reduces redundancy and induces diversity in the learning of the patterns. \\
Figure \ref{fig:cifar_pixel_interpret} visualizes how the image decomposes into patches and how patterns in each patch are represented by multiple clauses. The figure shows how patterns are identified in an image by the clauses in terms of the pixels (literals in the clause). In this case, each clause is comprised of an input of $64$ pixels in $8\times8$ patch form. The clauses are formed using these pixels as literals. The pixels are simply joined together by logical \emph{NOT} $(\neg)$ and \emph{AND} $(\land)$ operators to form Boolean propositional logic expressions. The figure further depicts three patches that have been extracted from different locations in the image. The clause expressions shown in Figure \ref{fig:cifar_pixel_interpret} are the ones that are activated in that particular patch for a particular class (\emph{ship} class in this case). Since the clauses are logical AND-rules, localized within the patches, they support pixel-level interpretation. The learning capability of drop clause TM is evident from this example, that how it is able to capture such patters from images with binary pixels, with just 1-bit per channel.
\vspace{-0.5cm}
\vspace{-1cm}

\section{Enhanced Robustness}
In this section, we show empirically that introduction of drop clause makes TM more robust towards input perturbations during testing. We evaluate robustness on both sentiment analysis and image classification\footnote{Note that we do not test for adversarial robustness. We only test the robustness of our models against input perturbations and corruptions.}.

\subsection{Natural Language Sentiment Analysis}
We select the MR dataset to test the robustness of TM on natural language sentiment analysis. We train our models on the standard training data and perturb the testing data to evaluate robustness towards out of distribution changes. We monitor the decrease in testing accuracy with and without perturbations. During testing, an input instance (sentence) is selected for perturbation with probability $0.5$. Then, a word is chosen from this sentence randomly with uniform probability. The selected word is swapped with the most similar word according to the cosine similarity between their respective Glove embedding vectors. Swapping with the most similar word results in changing the binary feature representation by a hamming distance of $2$, while keeping the meaning and sentiment of the sentence unchanged. The following is an MR example demonstrating this strategy: \emph{\textcolor{red}{Magnificent} drama well worth watching} $\rightarrow$ \emph{\textcolor{green}{Splendid} drama well worth watching}. The word \emph{Magnificent} is changed to its synonym \emph{Splendid}. Table \ref{tab:nlpr} shows the comparison between TM and drop clause TM on MR with and without test data perturbations. The difference between the test accuracy on the dataset with and without perturbations is $2.02\%$ and $1.45\%$ for the standard TM and drop clause TM respectively. In addition, drop clause makes TM more robust compared to BiLSTM, where the accuracy drops by $1.83\%$. The increase in TM robustness could be due to the fact that since clauses are stochastically dropped they seem to be less prone to pick up irrelevant or noisy patterns, making them more robust.
\vspace{-0.3cm}
\begin{table}[h]
\caption{Robustness Comparison on MR dataset}
\label{tab:nlpr}
\begin{center}
\begin{small}
\begin{sc}
\adjustbox{width=0.48\textwidth}{%
\begin{tabular}{lccc}
\toprule
 & TM & TM (Drop Clause) & BiLSTM \\
    \midrule
    MR & 75.14 & 78.67 &  77.68 \\
    MR (perturb) &  73.12 & 77.12 & 75.85  \\
\bottomrule
\end{tabular}}
\end{sc}
\end{small}
\end{center}
\end{table}
\vspace{-0.75cm}
\begin{table}[h]
\caption{Robustness Comparison on MNIST-C dataset}
\label{tab:mnistc}
\begin{center}
\begin{small}
\begin{sc}
\begin{tabular}{lccc}
\toprule
 & CTM & CTM (Drop Clause) & BNN \\
    \midrule
    MNIST & 99.3 & 99.45 &  98.25 \\
    MNIST-C &  88.12 & 91.75 &  85.5 \\
\bottomrule
\end{tabular}
\end{sc}
\end{small}
\end{center}
\vskip -0.25in
\end{table}

\subsection{Image Classification}
We compare the robustness of TM and drop clause TM using the MNIST-C dataset \cite{mnistc}, which is a \emph{corrupted} version of MNIST. It consists of $15$ corruptions such as Gaussian blur, scale, rotate, shear, impulse noise, canny edges, fog etc. We train our models on the standard MNIST training data without any augmentations. During testing, we select whether to use non-corrupted or corrupted test image with probability $0.5$. We then select one of the $15$ corruptions to apply on the test image with uniform probability. Drop clause enhances the robustness of CTM, as can be seen in Table \ref{tab:mnistc}. There is a drop of $11.18\%$ for the standard CTM whereas the drop in accuracy is reduced to $7.7\%$ with drop clause. Also, from Table \ref{tab:mnistc}, we observe that both the TM versions are more robust than binarized neural networks (BNN)\footnote{The MNIST-MLP architecture from \cite{bnncompare} is used here.}, whose accuracy drops drops by $12.75\%$.
\vspace{-0.1cm}
\section{Discussion}
\vspace{-0.1cm}
To summarize our results, drop clause improves the performance of the TM, enhancing the advantages TMs have over neural networks and traditional machine learning techniques when it comes to computational complexity, memory consumption, training and inference time, and perhaps most importantly, interpretability. The computational advantages make TMs suitable for federated learning and deployment on edge devices. On the other hand, the TM still achieves lower accuracy on the datasets we use here when compared to some deep neural network models. We conjecture that the accuracy gap could be caused by the Booleanization information loss, e.g., going from 3x8 bits pixel values to 3 Boolean values (3 bits). Thresholding the 3 color channels separately may also be sub-optimal because the combinations of the three colour channels produce different colours and textures. Another challenge arises in NLP tasks. Here, pretrained models have been dominating, leveraging unlabelled data. Important models include word2vec, GloVe, BERT, and GPT. TMs require Boolean input, making it difficult to leverage existing high dimensional embeddings, impeding performance, with the exception of enhancing the input with GloVe-derived synonyms \cite{rohan1}. Also, the pixel-level interpretability on image classification can be difficult for humans to interpret as the Boolean expressions for clauses capturing patterns can be quite long containing numerous pixels, especially for large images.

\section{Conclusion}
\vspace{-0.1cm}
In this paper, we propose \emph{drop clause} as a technique to improve the generalization ability of the TM. Drop clause enhances stochasticity during training, to provide more diverse patterns. As a result, TM performance is significantly boosted which is empirically shown on a diverse collection of datasets, both in terms of accuracy and training time. We further show how drop clause improves the pattern recognition capabilities of TM, simultaneously improving pattern interpretability. When comparing our model with state-of-the-art deep learning   models, we observe competitive accuracy levels. Since TM is more akin to traditional machine learning techniques, we also compare it with a selection of those, reporting superior accuracy results for the TM. We further showcase pixel-level interpretability on the CIFAR-10 dataset and word-level interpretability on the MR dataset. We finally establish that drop clause improves the robustness of the TM towards data corruptions and perturbations during testing.

\bibliography{example_paper}
\bibliographystyle{icml2022}

\newpage
\appendix
\onecolumn
\section{Appendix}
\subsection{Tsetlin Machine}
\begin{figure}[h]
  \begin{center}
  \includegraphics[width=\linewidth]{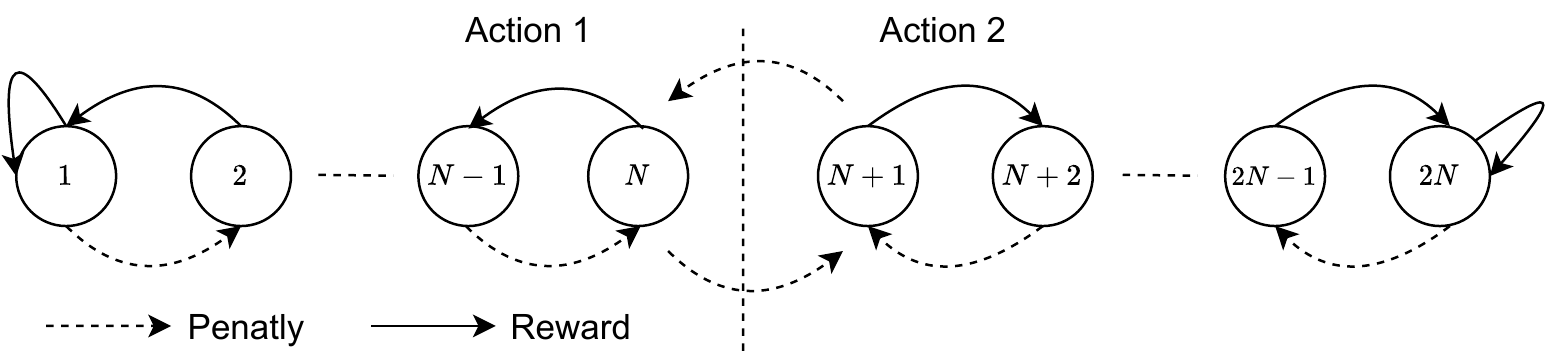}
  \caption{A two-action Tsetlin Automaton with $2N$ states.}\label{figTA}
  \end{center}
\end{figure}

\begin{figure}[ht]
\begin{center}
\centerline{\includegraphics[width=\columnwidth]{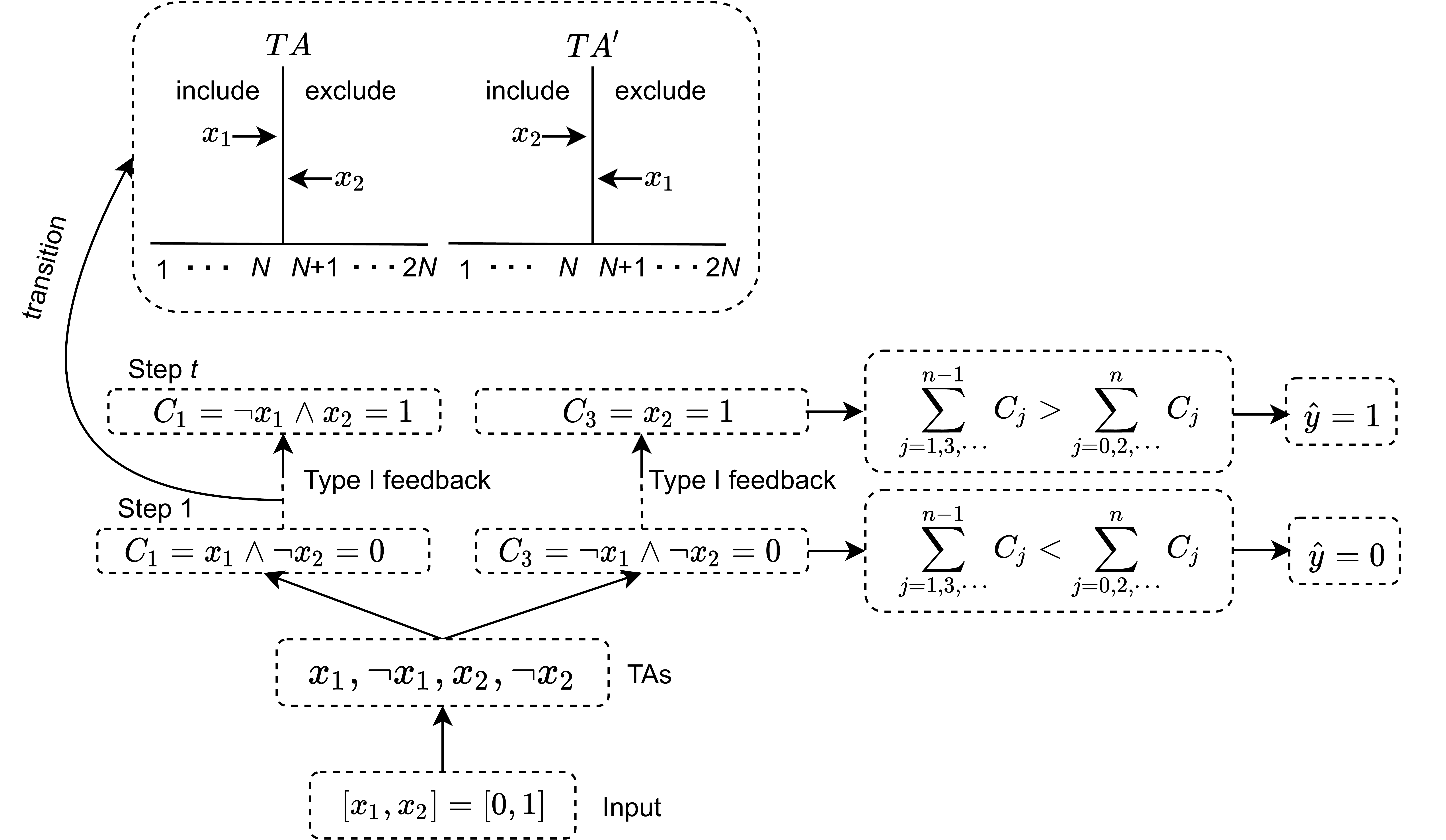}}
\caption{TM learning dynamics for an XOR-gate training sample, with input ($x_1=0, x_2=1$) and output target $y=1$.}
\label{figure:tm_architecture_basic}
\end{center}
\end{figure}

\paragraph{Structure.} A TM in its simplest form takes a feature vector $\mathbf{x} = [x_1, x_2, \ldots, x_o] \in \{0,1\}^o$ of $o$ propositional values as input and assigns the vector a class $\hat{y} \in \{0,1\}$. To minimize classification error, the TM produces $n$ self-contained patterns. In brief, the input vector $\mathbf{x}$ provides the literal set $L = \{l_1, l_2, \ldots, l_{2o}\} = \{x_1, x_2, \ldots, x_{o}, \lnot x_1, \lnot x_2, \ldots, \lnot x_o\}$, consisting of the input features and their negations. By selecting subsets $L_j \subseteq L$ of the literals, the TM can build arbitrarily complex patterns, ANDing the selected literals to form conjunctive clauses:
\begin{equation}
C_j(\mathbf{x})= \bigwedge_{l_k \in L_j} l_k.
\end{equation}
Above, $j \in \{1, 2, \ldots, n\}$ refers to a particular clause $C_j$ and $k \in \{1, 2, \ldots, 2o\}$ refers to a particular literal $l_k$. As an example, the clause $C_j(\mathbf{x}) = x_1 \land \lnot x_2$ consists of the literals $L_j = \{x_1, \lnot x_2\}$ and evaluates to $1$ when $x_1=1$ and $x_2=0$. 

The TM assigns one TA per literal $l_k$ per clause $C_j$ to build the clauses. The TA assigned to literal $l_k$ of clause $C_j$ decides whether $l_k$ is \emph{Excluded} or \emph{Included} in $C_j$. Figure~\ref{figTA} depicts a two-action TA with $2N$ states.  For states $1$ to $N$, the TA performs action \emph{Exclude} (Action 1), while for states $N + 1$ to $2N$ it performs action \emph{Include} (Action 2). As feedback to the action performed, the environment responds with either a Reward or a Penalty. If the TA receives a Reward, it moves deeper into the side of the action. If it receives a Penalty, it moves towards the middle and eventually switches action.

With $n$ clauses and $2o$ literals, we get $n\times2o$ TAs. We organize the states of these in a $n\times2o$ matrix $A = [a_k^j] \in \{1, 2, \ldots, 2N\}^{n\times2o}$. We will use the function $g(\cdot)$ to map the automaton state $a_k^j$ to Action $0$ (\emph{Exclude}) for states $1$ to $N$ and to Action $1$ (\emph{Include}) for states $N+1$ to~$2N$: $g(a_k^j) = a_k^j > N$.

We can connect the states $a_k^j$ of the TAs assigned to clause $C_j$ with its composition as follows:
\begin{equation}
C_j(\mathbf{x}) = \bigwedge_{l_k \in L_j} l_k = \bigwedge_{k=1}^{2o} \left[g(a_k^j) \Rightarrow l_k\right].
\end{equation}
Here, $l_k$ is one of the literals and $a_k^j$ is the state of its TA in clause $C_j$. The logical \emph{imply} operator~$\Rightarrow$ implements the \emph{Exclude}/\emph{Include} action. That is, the 
\emph{imply} operator is always $1$ if $g(a_k^j)=0$ (\emph{Exclude}), while if $g(a_k^j)=1$ (\emph{Include}) the truth value is decided by the truth value of the literal.

\paragraph{Classification.} Classification is performed as a majority vote. The odd-numbered half of the clauses vote for class $\hat{y} = 0$ and the even-numbered half vote for $\hat{y} = 1$:
\begin{equation}
    \hat{y} = 0 \le \sum_{j=1,3,\ldots}^{n-1} \bigwedge_{k=1}^{2o} \left[g(a_k^j) \Rightarrow l_k\right] - \sum_{j=2,4,\ldots}^{n} \bigwedge_{k=1}^{2o} \left[g(a_k^j) \Rightarrow l_k\right]. \label{eqn:prediction}
\end{equation}
As such, the odd-numbered clauses have positive polarity, while the even-numbered ones have negative polarity. As an example, consider the input vector $\mathbf{x} = [0, 1]$ in the lower part of Figure \ref{figure:tm_architecture_basic}. The figure depicts two clauses of positive polarity, $C_1(\mathbf{x}) = x_1 \land \lnot x_2$ and $C_3(\mathbf{x}) = \lnot x_1 \land \lnot x_2$ (the negative polarity clauses are not shown). Both of the clauses evaluate to zero, leading to class prediction $\hat{y} = 0$.

\begin{table}[t]
\centering
\vskip 0.15in
\begin{center}
\begin{small}
\begin{sc}
\begin{tabular}{l|l|l|l}
    \hline
    \multirow{2}{*}{Input}&Clause & \ \ \ \ \ \ \ 1 & \ \ \ \ \ \ \ 0 \\
    &{Literal} &\ \ 1 \ \ \ \ \ \ 0 &\ \ 1 \ \ \ \ \ \ 0 \\
    \hline
    \multirow{2}{*}{Include Literal}&P(Reward)&$\frac{s-1}{s}$\ \ \ NA & \ \ 0 \ \ \ \ \ \ 0\\ [1mm]
    &P(Inaction)&$\ \ \frac{1}{s}$\ \ \ \ \ NA &$\frac{s-1}{s}$ \ $\frac{s-1}{s}$ \\ [1mm]
    &P(Penalty)& \ \ 0 \ \ \ \ \ NA& $\ \ \frac{1}{s}$ \ \ \ \ \  $\frac{1}{s}$ \\ [1mm]
    \hline
    \multirow{2}{*}{Exclude Literal}&P(Reward)& \ \ 0 \ \ \ \ \ \ $\frac{1}{s}$ & $\ \ \frac{1}{s}$ \ \ \ \ \
    $\frac{1}{s}$ \\ [1mm]
    &P(Inaction)&$ \ \ \frac{1}{s}$\ \ \ \ $\frac{s-1}{s}$  &$\frac{s-1}{s}$ \ $\frac{s-1}{s}$ \\ [1mm]
    &P(Penalty)&$\frac{s-1}{s}$ \ \ \ \ 0& \ \ 0 \ \ \ \ \ \ 0 \\ [1mm]
    \hline
\end{tabular}
\end{sc}
\end{small}
\end{center}
\caption{Type I Feedback}
\label{table:type_i}
\end{table}

\begin{table}[t]
\centering
\vskip 0.15in
\begin{center}
\begin{small}
\begin{sc}
\begin{tabular}{l|l|l|l}
    \hline
    \multirow{2}{*}{Input}&Clause & \ \ \ \ \ \ \ 1 & \ \ \ \ \ \ \ 0 \\
    &{Literal} &\ \ 1 \ \ \ \ \ \ 0 &\ \ 1 \ \ \ \ \ \ 0 \\
    \hline
    \multirow{2}{*}{Include Literal}&P(Reward)&\ \ 0 \ \ \ NA & \ \ 0 \ \ \ \ \ \ 0\\[1mm]
    &P(Inaction)&1.0 \ \  NA &  1.0 \ \ \ 1.0 \\[1mm]
    &P(Penalty)&\ \ 0 \ \ \ NA & \ \ 0 \ \ \ \ \ \ 0\\[1mm]
    \hline
    \multirow{2}{*}{Exclude Literal}&P(Reward)&\ \ 0 \ \ \ \ 0 & \ \ 0 \ \ \ \ \ \ 0\\[1mm]
    &P(Inaction)&1.0 \ \ \ 0 &  1.0 \ \ \ 1.0 \\[1mm]
    &P(Penalty)&\ \ 0 \ \  1.0 & \ \ 0 \ \ \ \ \ \ 0\\[1mm]
    \hline
\end{tabular}
\end{sc}
\end{small}
\end{center}
\caption{Type II Feedback}
\label{table:type_ii}
\end{table}

\paragraph{Learning.} The upper part of Figure \ref{figure:tm_architecture_basic} illustrates learning. A TM learns online, processing one training example $(\mathbf{x}, y)$ at a time. Based on $(\mathbf{x}, y)$, the TM rewards and penalizes its TAs, which amounts to incrementing and decrementing their states. There are two kinds of feedback: Type I Feedback produces frequent patterns and Type II Feedback increases the discrimination power of the patterns.

Type I feedback is given stochastically to clauses with positive polarity when $y=1$  and to clauses with negative polarity when $y=0$. Conversely, Type II Feedback is given stochastically to clauses with positive polarity when $y=0$ and to clauses with negative polarity when $y=1$. The probability of a clause being updated is based on the vote sum $v$: $v = \sum_{j=1,3,\ldots}^{n-1} \bigwedge_{k=1}^{2o} \left[g(a_k^j) \Rightarrow l_k\right] - \sum_{j=2,4,\ldots}^{n} \bigwedge_{k=1}^{2o} \left[g(a_k^j) \Rightarrow l_k\right]$. The voting error is calculated as:
\begin{equation}
\epsilon = \begin{cases}
T-v& y=1\\
T+v& y=0.
\end{cases}
\end{equation}
Here, $T$ is a user-configurable voting margin yielding an ensemble effect. The probability of updating each clause is $P(\mathrm{Feedback}) = \frac{\epsilon}{2T}$.

After random sampling from $P(\mathrm{Feedback})$ has decided which clauses to update, the following TA state updates can be formulated as matrix additions, subdividing Type I Feedback into feedback Type Ia and Type Ib:
\begin{equation}
    A^*_{t+1} = A_t + F^{\mathit{II}} + F^{Ia} - F^{Ib}.
    \label{eqn:learning_step_1}
\end{equation}
Here, $A_t = [a^j_k] \in \{1, 2, \ldots, 2N\}^{n \times 2o}$ contains the states of the TAs at time step $t$ and $A^*_{t+1}$ contains the updated state for time step $t+1$ (before clipping). The matrices $F^{\mathit{Ia}} \in \{0,1\}^{n \times 2o}$ and $F^{\mathit{Ib}} \in \{0,1\}^{n \times 2o}$ contains Type I Feedback. A zero-element means no feedback and a one-element means feedback. As shown in Table \ref{table:type_i}, two rules govern Type I feedback:
\begin{itemize}
    \item \textbf{Type Ia Feedback} is given with probability $\frac{s-1}{s}$ whenever both clause and literal are $1$-valued.\footnote{Note that the probability $\frac{s-1}{s}$ is replaced by $1$ when boosting true positives.} It penalizes \emph{Exclude} actions and rewards \emph{Include} actions. The purpose is to remember and refine the patterns manifested in the current input $\mathbf{x}$. This is achieved by increasing selected TA states. The user-configurable parameter $s$ controls pattern frequency, i.e., a higher $s$ produces less frequent patterns.
    \item \textbf{Type Ib Feedback} is given with probability $\frac{1}{s}$ whenever either clause or literal is $0$-valued. This feedback rewards \emph{Exclude} actions and penalizes \emph{Include} actions to coarsen patterns, combating overfitting. Thus, the selected TA states are decreased.
\end{itemize}

The matrix $F^{\mathit{II}} \in \{0, 1\}^{n \times 2o}$ contains Type II Feedback to the TAs, given per Table \ref{table:type_ii}.
\begin{itemize}
\item \textbf{Type II Feedback} penalizes \emph{Exclude} actions to make the clauses more discriminative, combating false positives. That is, if the literal is $0$-valued and the clause is $1$-valued, TA states below $N+1$ are increased. Eventually the clause becomes $0$-valued for that particular input, upon inclusion of the $0$-valued literal.
\end{itemize}

The final updating step for training example  $(\mathbf{x}, y)$ is to clip the state values to make sure that they stay within value $1$ and $2N$:
\begin{equation}
    A_{t+1} = \mathit{clip}\left(A^*_{t+1}, 1, 2N\right). \label{eqn:learning_step_2}
\end{equation}

For example, both of the clauses in Figure \ref{figure:tm_architecture_basic} receives Type I Feedback over several training examples, making them resemble the input associated with $y=1$.

\begin{figure}[!h]
\centering
\begin{subfigure}[t]{.49\textwidth}
\includegraphics[width=\linewidth]{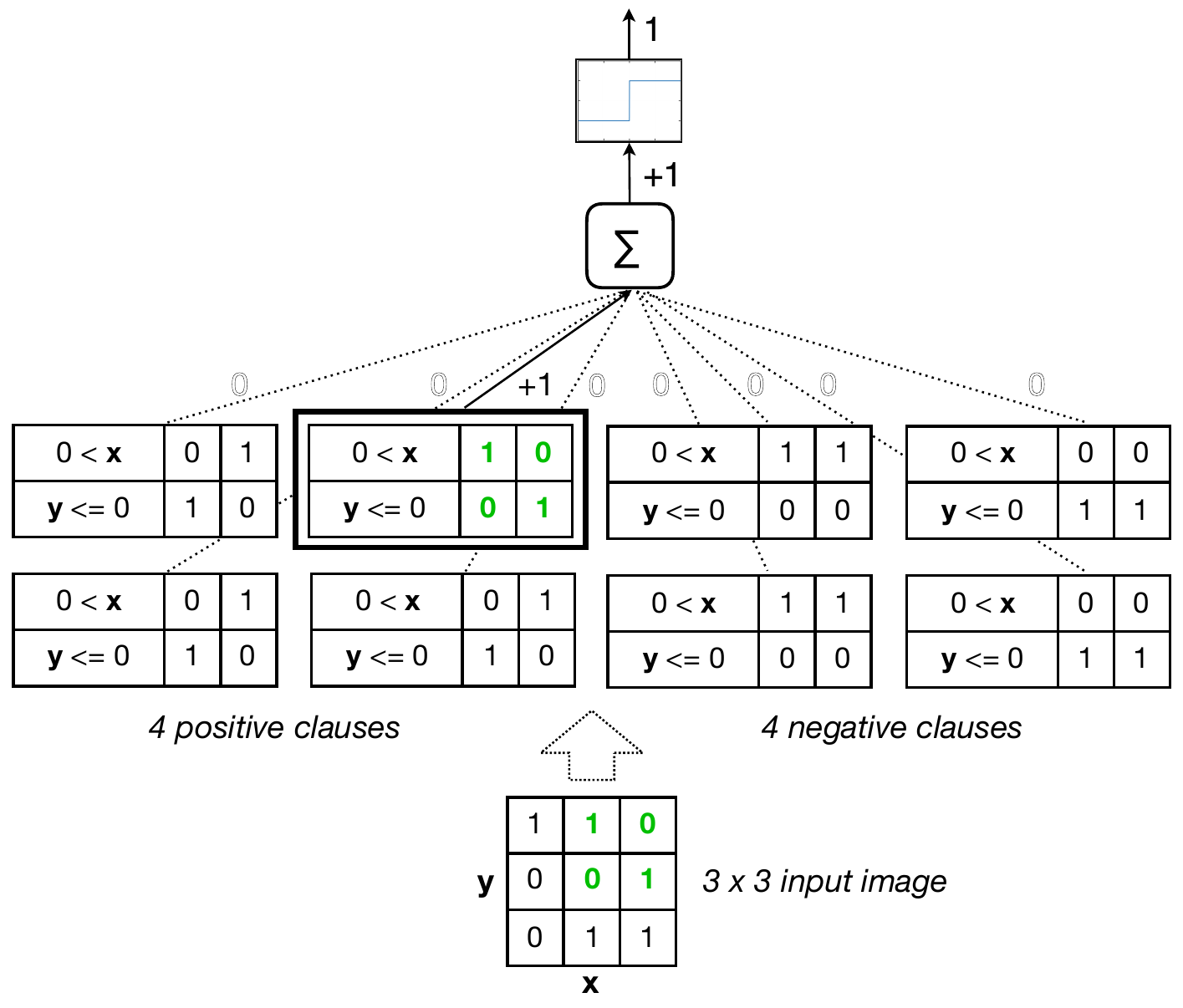}
\caption{}
\label{figure:inference_example}
\end{subfigure}
\hspace{0mm}
\begin{subfigure}[t]{.49\textwidth}
\includegraphics[width=\linewidth]{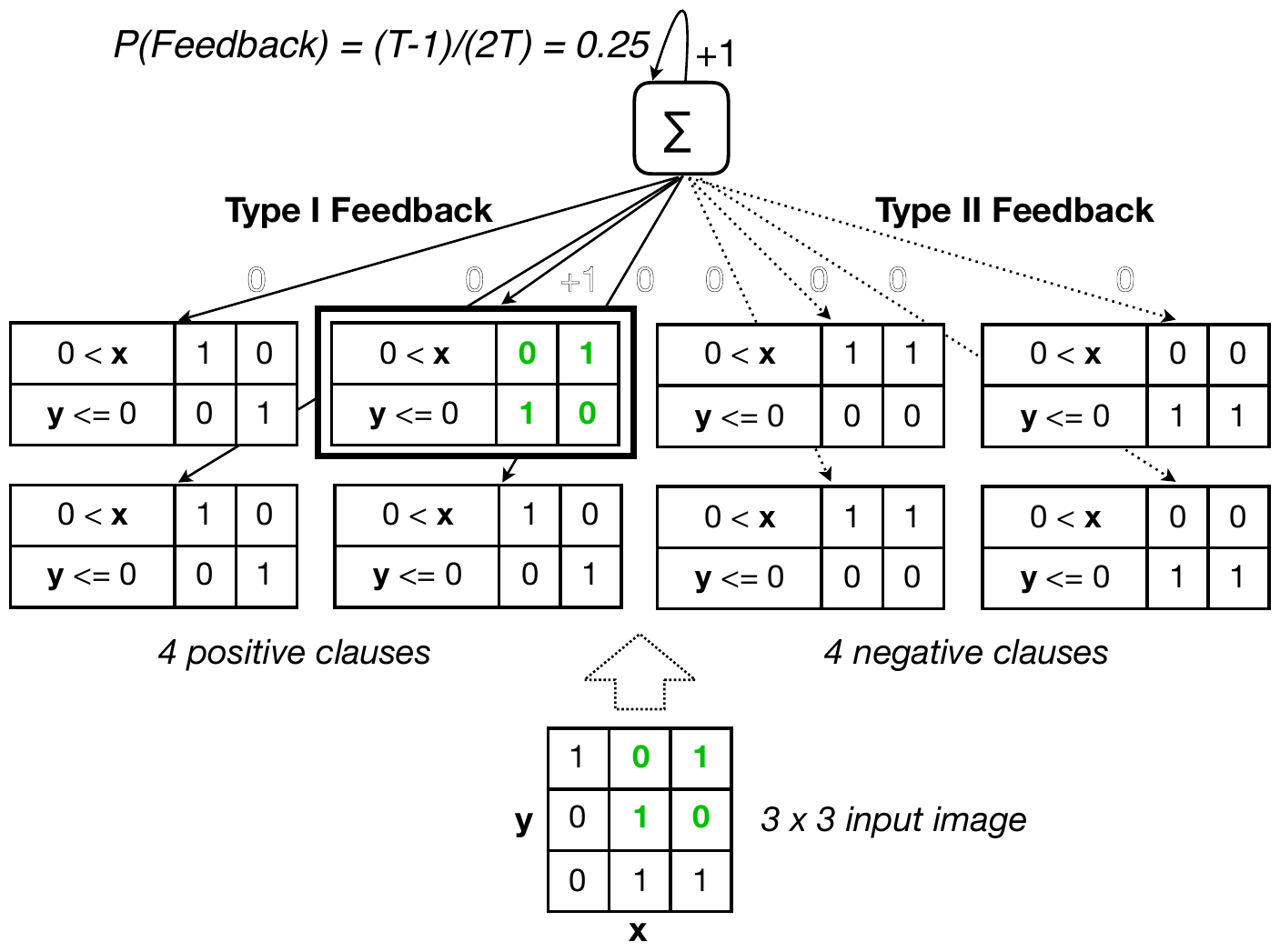}
\caption{}
\label{figure:learning_example}
\end{subfigure}
\caption{Example of inference (a) and learning (b) for the Noisy 2D XOR Problem.}
\end{figure}

\begin{figure}[!h]
\centering
\begin{subfigure}[t]{.49\textwidth}
\includegraphics[width=\linewidth]{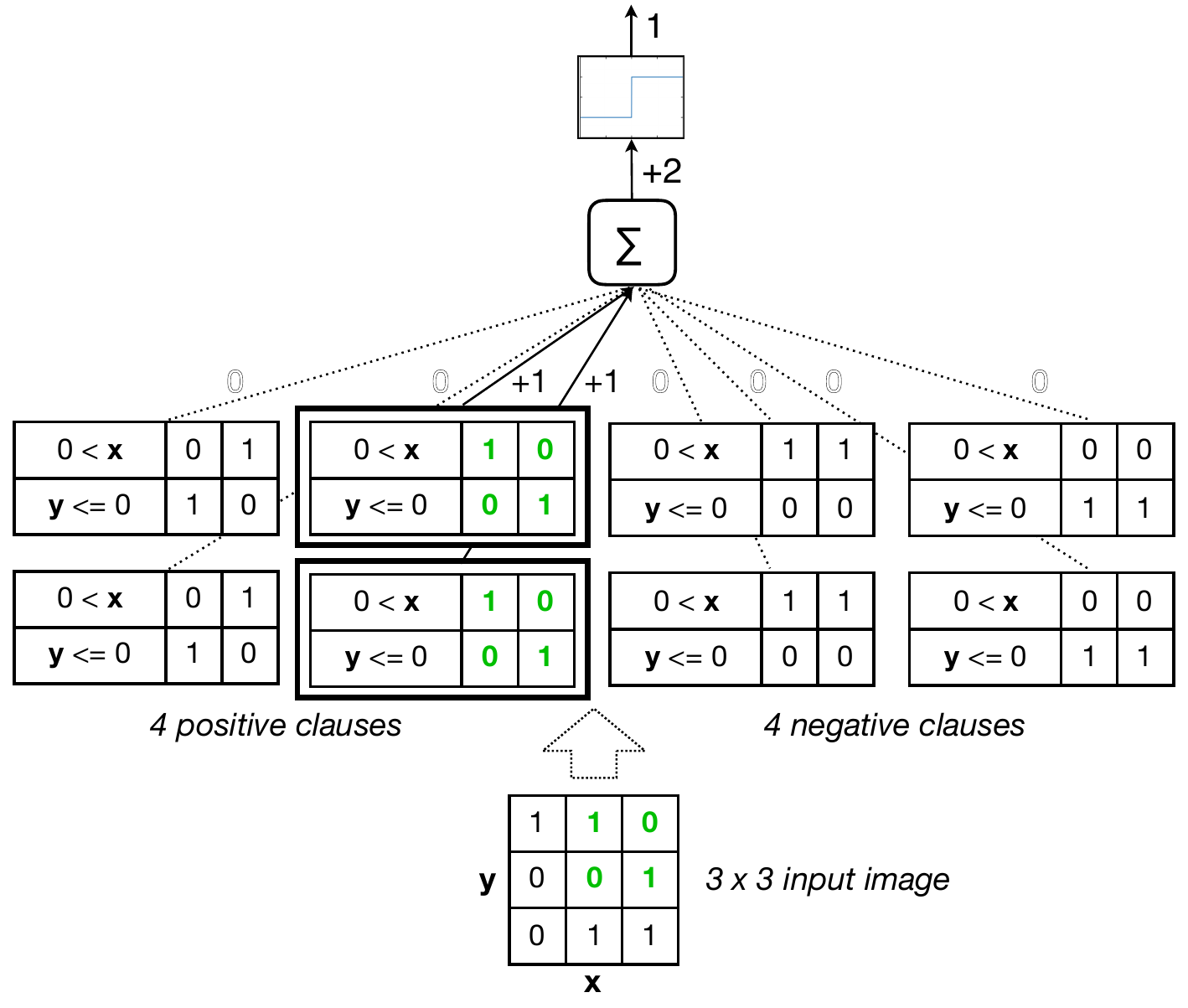}
\caption{}
\label{figure:goal_state}
\end{subfigure}
\hspace{0mm}
\begin{subfigure}[t]{.4\textwidth}
\includegraphics[width=\linewidth]{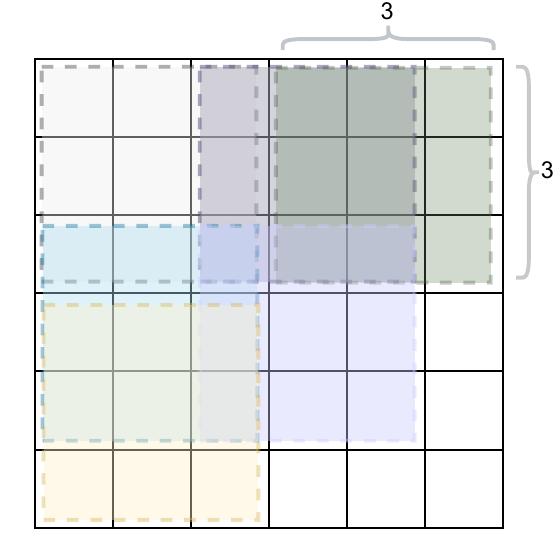}
\caption{}
\label{figure:convolution}
\end{subfigure}
\caption{(a) Goal state for the Noisy 2D XOR Problem. (b) Illustration of image, filter and patches.}\label{figure:example}
\end{figure}

\subsection{Convolutional Tsetlin Machine}
Consider a set of images $\mathcal{X}=\{\mathbf{x}_e | 1 \le e \le E\}$, where $e$ is the index of the images. Each image is of size $d_x \times d_y$ and consists of $d_z$ binary layers, illustrated in Figure \ref{figure:example}b. A classic TM models such an image with an input vector $\mathbf{x} = [x_k] \in \{0, 1\}^{d_x \times d_y \times d_z}$ that contains $d_x \times d_y \times d_z$ input features. Accordingly, each clause is composed from $d_x \times d_y \times d_z \times 2$ literals.

\paragraph{Structure.} The CTM (CTM) \cite{granmo2019convtsetlin} performs a convolution over the input image $\mathbf{x}$, dividing it into patches with spatial dimensions $d_w \times d_w$. That is, the input vector $\mathbf{x} = [x_k] \in \{0, 1\}^{d_x \times d_y \times d_z}$ produces $B = \left( \ceil*{\frac{d_x -  d_w}{q}}+1\right) \times \left(\ceil*{\frac{d_y - d_w}{q}}+1\right)$ patches, with $q$ being the step size of the convolution. For instance, Figure \ref{figure:example}b illustrates $B = (6-3 + 1)\times(6-3+1)=16$ patches of size $3\times3$, assuming step size $q=1$.

Each patch $b \in \{1, 2, \ldots, B\}$, in turn, yields an input vector $\mathbf{x}^b = [x_k^b] \in \{0, 1\}^{d_w \times d_w \times d_z}$ with a corresponding literal vector $\mathbf{l}^b = [l_k^b] \in \{0, 1\}^{d_w \times d_w \times d_z \times 2}$. The CTM becomes location aware \cite{Liu2018} by augmenting each patch input vector $\mathbf{x}^b$ with the coordinates of $\mathbf{x}^b$ within $\mathbf{x}$, using threshold-based encoding \cite{abeyrathna2020nonlinear}.

\paragraph{Classification.} The CTM is based on the classic TM procedure for classification (Eqn. \ref{eqn:prediction}). However, we now have $B$ input vectors $\mathbf{x}^b$ per image rather than a single input vector $\mathbf{x}$. The convolution is performed by evaluating each clause $C_j$ on each input vector $\mathbf{x}^b$, i.e., calculating $\bigwedge_{k=1}^{2o} \left[g(a_k^j) \Rightarrow l_k^b\right]$, and then ORing the evaluations per clause:
\begin{equation}
    \hat{y} = 0 \le \sum_{j=1,3,\ldots}^{n-1}  \bigvee_{b=1}^B \left[\bigwedge_{k=1}^{2o} \left[g(a_k^j) \Rightarrow l_k^b\right] \right] - \sum_{j=2,4,\ldots}^{n} \bigvee_{b=1}^B \left[ \bigwedge_{k=1}^{2o} \left[g(a_k^j) \Rightarrow l_k^b\right]\right]. \label{eqn:prediction_convolution}
\end{equation}
Figure \ref{figure:inference_example} provides an example where a $3\times3$ input image produces four $2\times2$ patches. The CTM has four clauses of positive polarity and four clauses of negative polarity. Only one of the clauses of positive polarity matches. This clause matches the upper left corner of the input image, hence evaluating to $1$. Accordingly, the net output sum is $+1$, yielding output $\hat{y}=1$.

\paragraph{Learning.} CTM learning leverages the TM learning procedure, per Eqn. \ref{eqn:learning_step_1} and Eqn. \ref{eqn:learning_step_2}. However, when giving Type Ia or Type II Feedback to each clause $C_j$, the CTM does not use the original input vector $\mathbf{x}$. Instead, it randomly selects one of the patch input vectors $\mathbf{x}^b$ that made the clause evaluate to $1$:
\begin{equation}
\mathbf{x}^b_j = \mathit{RandomChoice}\left(\left\{
 \mathbf{x}^b  \left|\bigwedge_{k=1}^{2o} \left[g(a_k^j) \Rightarrow l_k^b\right] = 1, 1 \le b \le B
 \right. \right\}\right).\end{equation}
For Type Ib Feedback, on the other hand, CTM follows the standard updating scheme. 

The reason for randomly selecting a patch input vector $\mathbf{x}^b$ is to have each clause extract a certain sub-pattern, and the randomness of the uniform distribution statistically spreads the clauses for different sub-patterns in the target image.

Figure~\ref{figure:learning_example} demonstrates a learning step. Only a single clause has recognized the input. Assuming a summation target (margin) of $T=2$ and net clause output sum $+1$ the probability of giving each clause feedback becomes $P(Feedback) = \frac{(2-1)}{2\cdot2} = 0.25$. Since the training example is $y=1$, the positive polarity clauses receives Type I Feedback with probability $0.25$, while the negative polarity clauses receive Type II feedback again with probability $0.25$. After several such updates, we have a more balanced representation of the input patterns in Figure~\ref{figure:goal_state}, with two clauses now recognizing the input.

\subsection{Convergence Insights}
It is analyzed in \cite{39} and \cite{jiao2021convergence} for the convergence properties of the vanilla TM for 1-bit case and XOR case respectively. Here we will briefly analyse the impact of the introduced randomized clause drop on the convergence. For the 1-bit case, the work in \cite{39} is to study the convergence feature of a TM with only one clause. When the randomized clause drop proposed is adopted, it will not influence the conclusions drawn in \cite{39} and the reasons are as follows.  The main difference between the proposed algorithm and the vanilla TM is that the clause in this work will not be updated upon each given training sample, but according to a pre-defined probability. In other words, the clause is updated based on a randomly down-sampled subset as a new training set compared with the original training data. Given infinite training data and ideally randomized down-sampling, the statistics of the samples in the subset is kept the same as the original training data set and the number of training samples is also sufficient. For this reason, the clause can still observe sufficient number of training samples and can also observe sufficient varieties of the training samples. In addition, the other updating rules are unchanged compared with the vanilla TM. Therefore, all conclusions in \cite{39} hold for the newly proposed algorithm. 
\par For the XOR case, there are two sub-patterns in the XOR operator and the TM is proven to be able to converge and learn both sub-patterns when threshold value $T$ is correctly configured \cite{jiao2021convergence}. According the the analysis in the previous paragraph, we understand that the training samples for the new algorithm is a down-sampled version of the original data set and the statistics of the samples is kept the same due to the ideal randomness. If the training samples with only one sub-pattern are given, the randomized down sampled data, given infinite time horizon, will still offer sufficient samples with the same probability distribution compared with the original training set. Therefore, Lemma 1 and Lemma 2 in \cite{jiao2021convergence} hold. In fact, when the training samples with both sub-patterns are given, Lemma 3 and Lemma 4 in \cite{jiao2021convergence} still hold. The reason is that ideal randomly down-sampling will not make the new training data biased in terms of the sub-patterns and thus will not change the nature of the recurrence for the clauses. Clearly, Lemma 5 in \cite{jiao2021convergence} is also true because the newly introduced process will not change the role of $T$. Theorem 2 in \cite{jiao2021convergence} is therefore self-evident. Indeed, the conclusion derived in \cite{jiao2021convergence} is applicable to the algorithm with drop clause.  

\par For the 1-bit case, as the analysis is to study the convergence feature of one clause, the randomized clause drop will not influence the conclusions drawn in \cite{39}. In this proposed algorithm, the clause will be updated randomly upon each given training sample. Consider infinite time horizon, the clause can still observe sufficient number of training samples. In addition, due to the randomness designed in this algorithm, the clause can also observe sufficient varieties of the training samples. In short, the clause is given a randomly down-sampled subset as the new training set from the original training data. Given infinite training data and ideally randomized down-sampling, the statistics of the training data set is kept the same and the number of training samples are sufficient. Therefore, all conclusions in \cite{39} for the newly proposed algorithm.      The main reason is that the randomized clause drop employed in this paper will only reduce the number of training samples that is to be observed by the TM. Given the sufficient randomized


\section{Related Work}
\label{sec:related}
There has been a surge of research in interpretable AI. Most of the studies focus on making neural networks more interpretable. Currently, state-of-the-art techniques use visual explanations and gradient interpretability. We briefly discuss selected related studies in this section. One approach is to compute the gradients of the class scores for an input image to visualize which parts of the input image impact the classification \cite{saliency}. Another approach to visualize class-specific score maps is introduced in \cite{interpretcnns}. The technique forces each filter in the top convolutional layers to learn a class-specific object by adding a modified mutual information loss. Training is end-to-end by adding the local filter loss to the task-specific loss. These steps produce a map of interest for the top convolutional layers. In \cite{patchnet}, the authors propose a two-stage approach. The first stage learns to estimate the conditional probability distribution of patches of pixels separately. The second stage, in turn, averages these estimates to obtain a global assessment. The Grad-CAM method, proposed in \cite{gradcam}, has been widely employed to visualize CNN outputs. Grad-CAM uses the gradients of a target class from the final convolution layer to produce a coarse localization map. The map highlights the regions in the image used by the neural network for predicting the concept. The work in  \cite{dissection} analyses how different layers, training conditions, model architecture, layer width, and accuracy impact visual interpretability. 

\par NLP model interpretation methods largely visualize attention weights for words. Example models include BERT and contextualized embedding \cite{bert,context}, which capture the semantic relatedness among words using context. However, the weights assigned by the attention vector to each input do not necessarily provide a faithful explanation of classification  \cite{attn,notxai}. Also, a general interpretability toolkit called InterpretML was developed \cite{interpretml} which explains linear machine learning algorithms and uses methods such as LIME to explain black-box models. However, the toolkit only outputs feature scores.

\par Overall, state-of-the-art methods use feature importance maps for interpretation, both for image classification- and NLP models. These maps only show where the neural network is ``looking''. Such an approach is not truly interpretable because one does not explain why attention is where it is.  Many researchers have attempted to replicate more human-level interpretation behavior in neural networks \cite{nlp1}, but have largely failed so far. In this paper, we improve the performance of the TM by introducing a novel method called drop clause. We also evaluate its convergence, accuracy, computation time, and interpretability. Additionally, we show how interpretable boolean expressions straightforwardly map to classification decisions, both in NLP and image classification.

\section{Enhanced Interpretability Examples}

\begin{figure}[b]
\centering
\begin{subfigure}{\textwidth}
\centering
    \includegraphics[width=\linewidth]{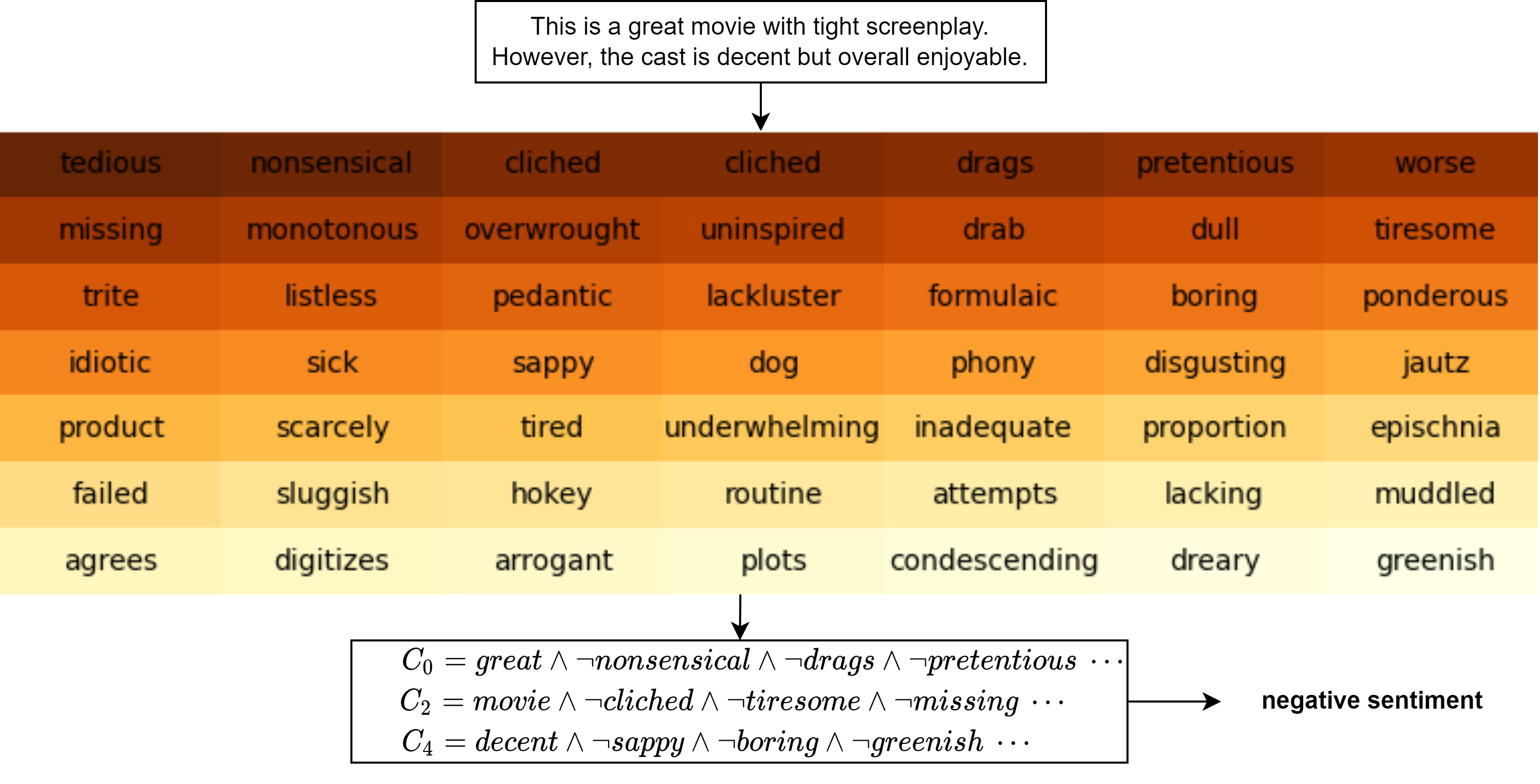}
    \caption{TM without drop clause}
    \vspace{0.5cm}
    \label{fig:inclu}
\end{subfigure}

\begin{subfigure}{\textwidth}
\centering
    \includegraphics[width=\linewidth]{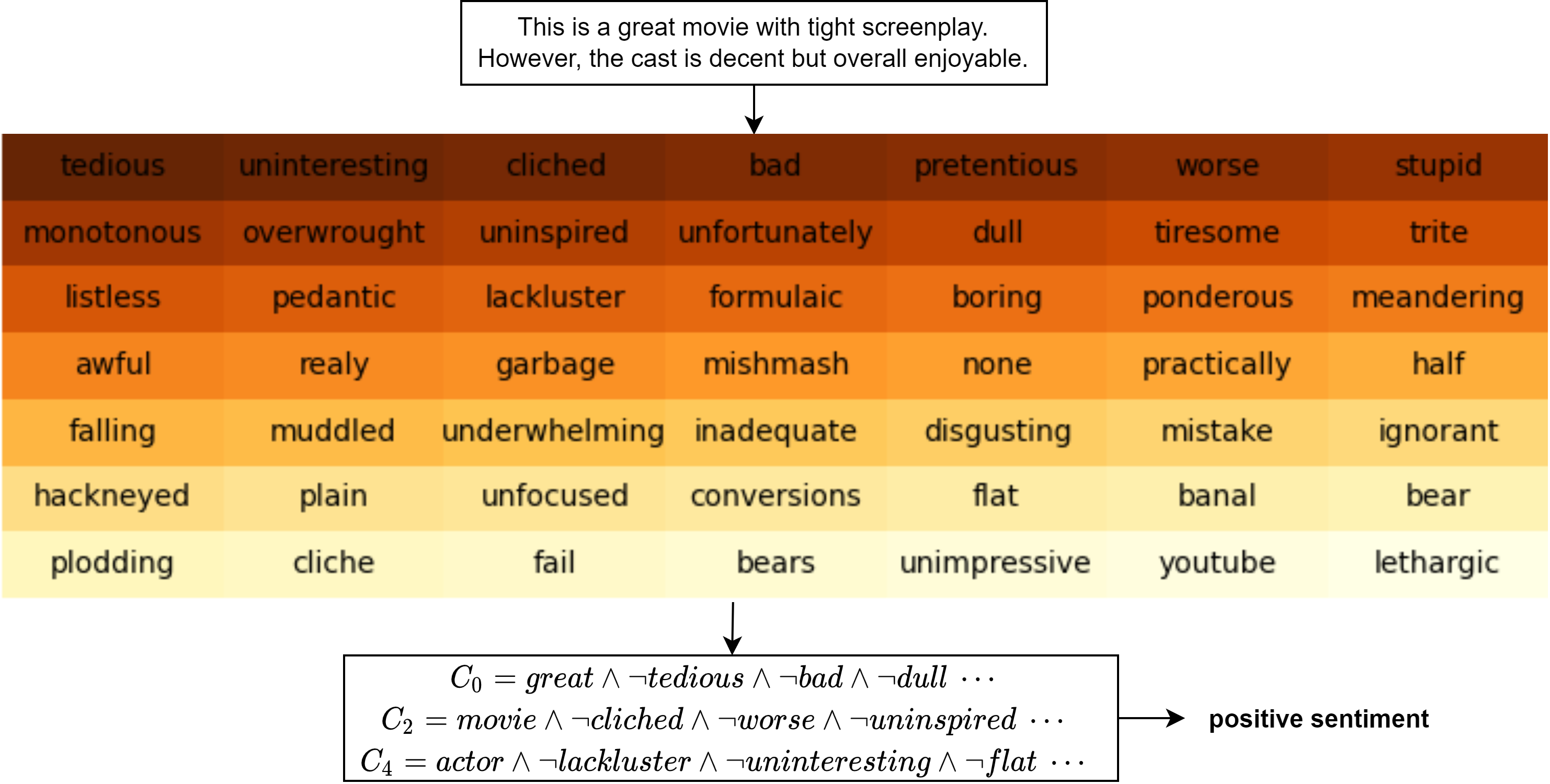}
    \caption{TM with drop clause $(p=0.75)$}
    \label{fig:deform}
\end{subfigure}
\caption{Natural Language Sentiment Analysis Interpretability: Example 1}
\label{fig:manmade}
\end{figure}

\begin{figure}[h]
\centering
\begin{subfigure}{\textwidth}
\centering
    \includegraphics[width=\linewidth]{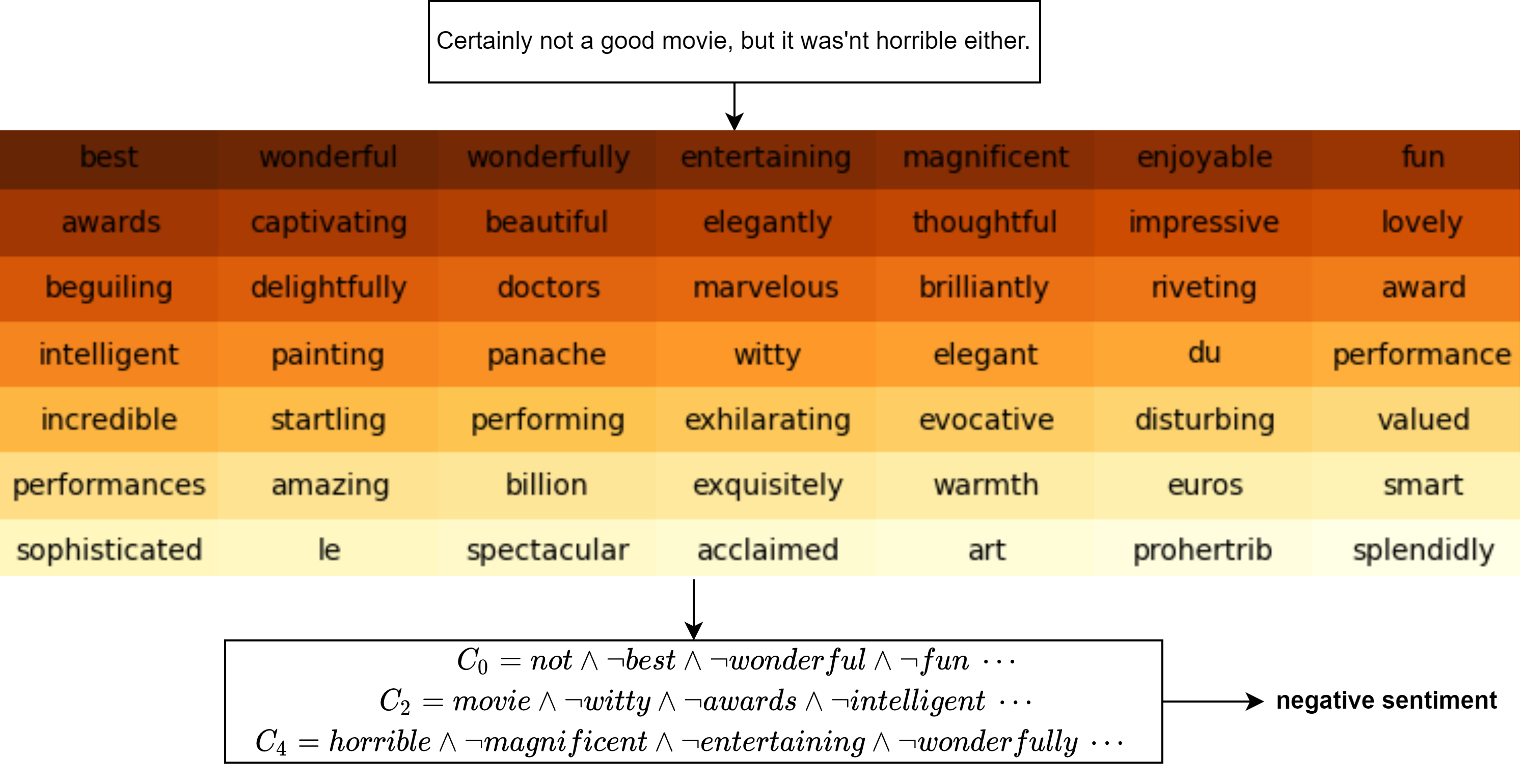}
    \caption{TM without drop clause}
    \vspace{0.5cm}
    \label{fig:inclu}
\end{subfigure}

\begin{subfigure}{\textwidth}
\centering
    \includegraphics[width=\linewidth]{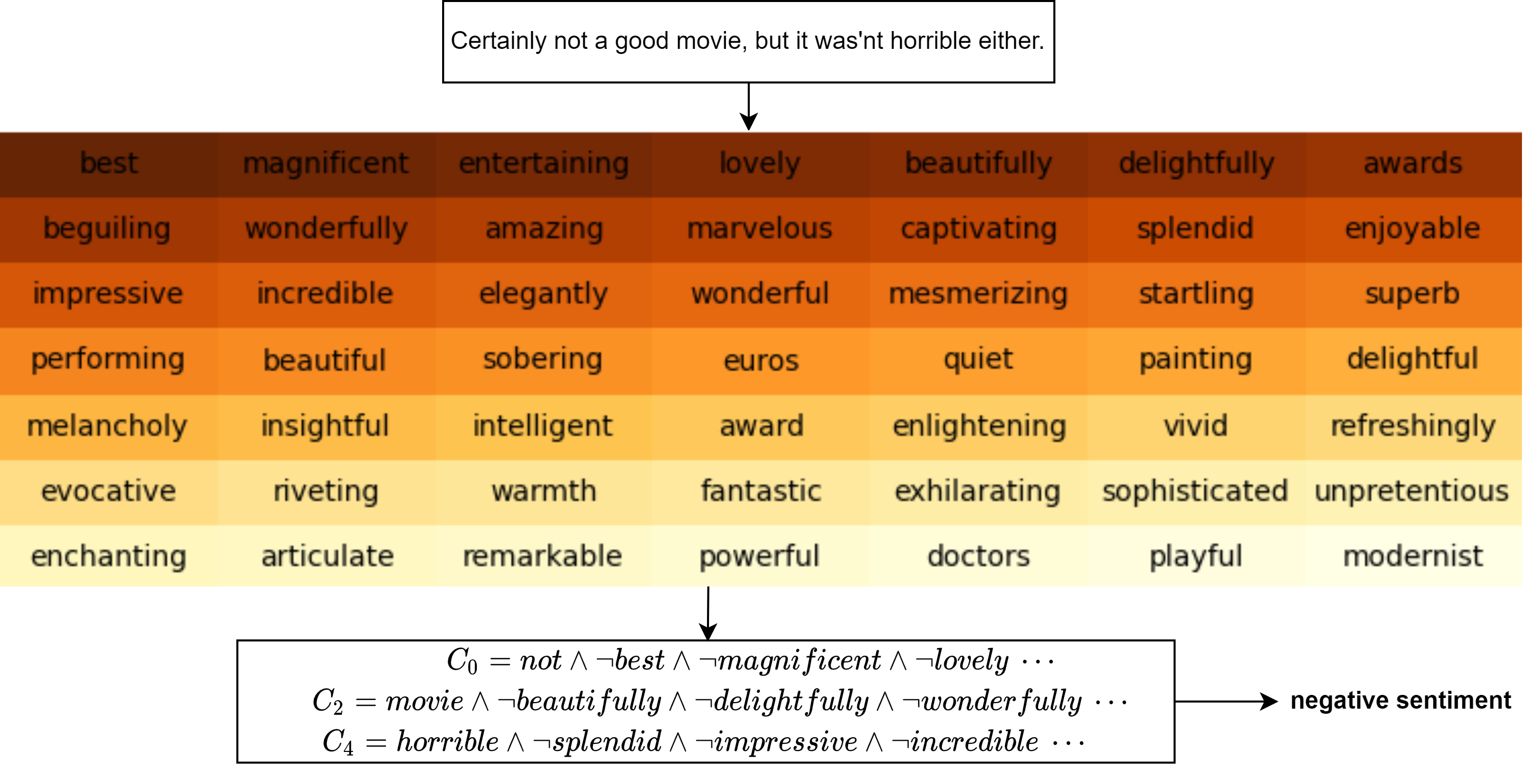}
    \caption{TM with drop clause $(p=0.75)$}
    \label{fig:deform}
\end{subfigure}
\caption{Natural Language Sentiment Analysis Interpretability: Example 2}
\label{fig:manmade}
\end{figure}

\begin{figure}[h]
		\centering
		\begin{tabular}{ccc}
			Image & TM Heatmap & TM $(p=0.5)$ Heatmap \\
			\subfloat{\includegraphics[width = 1.5in]{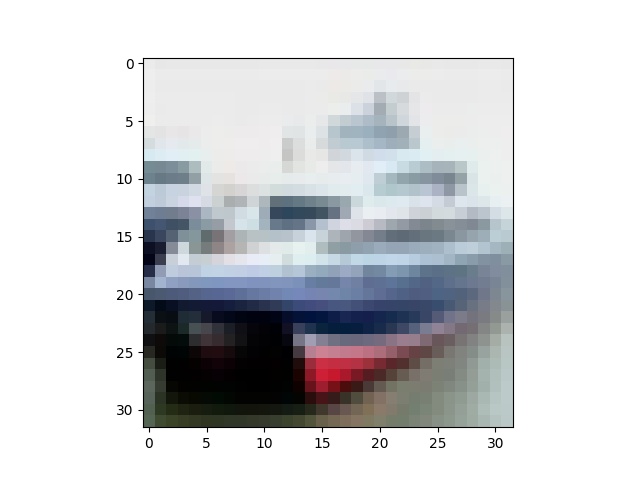}} &
			\subfloat{\includegraphics[width = 1.5in]{heatmap_0.png}} &
			\subfloat{\includegraphics[width = 1.5in]{heatmap_dc_0.png}} \\
			\subfloat{\includegraphics[width = 1.5in]{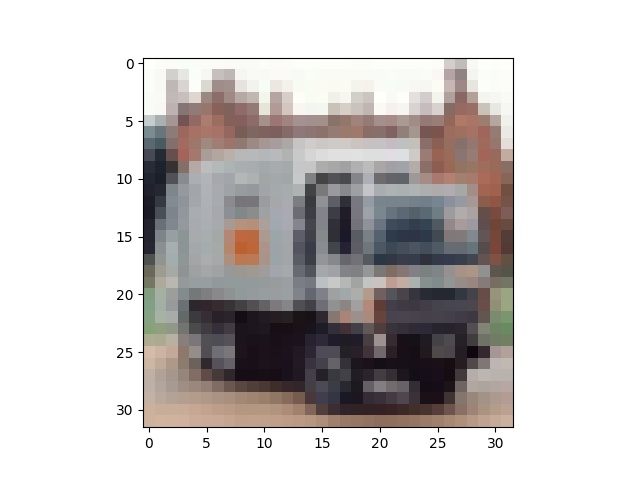}} &
			\subfloat{\includegraphics[width = 1.5in]{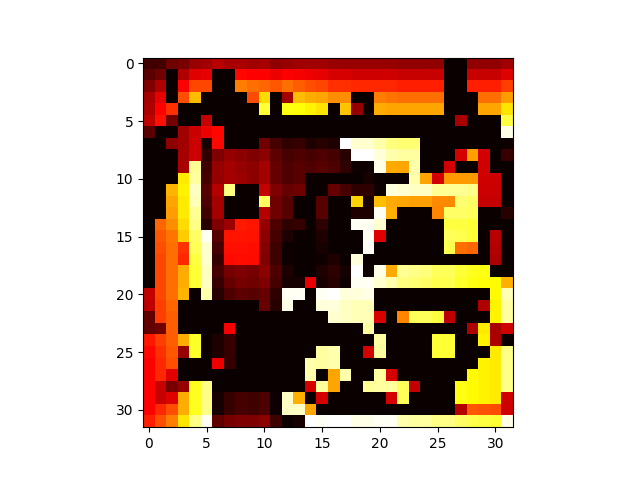}} &
			\subfloat{\includegraphics[width = 1.5in]{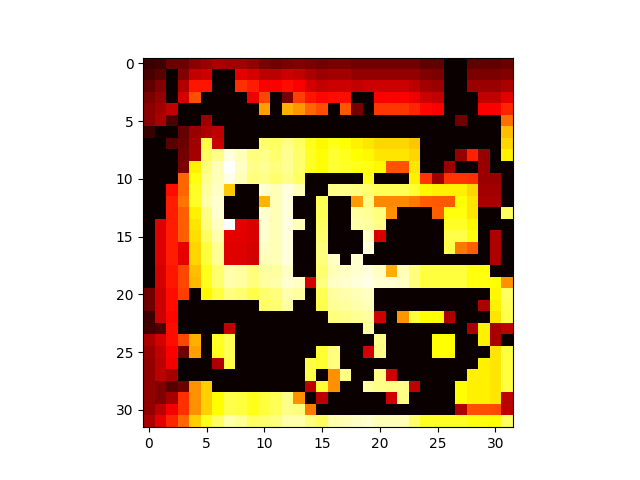}} \\
			\subfloat{\includegraphics[width = 1.5in]{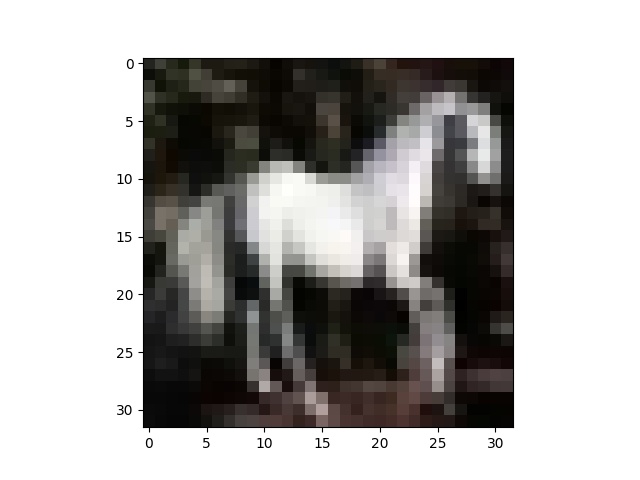}} &
			\subfloat{\includegraphics[width = 1.5in]{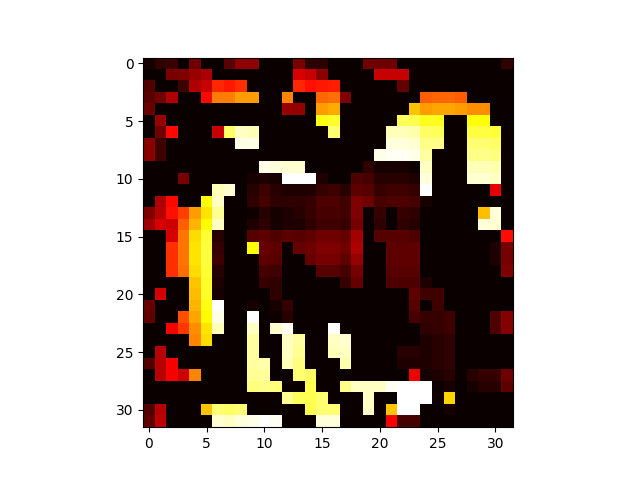}} &
			\subfloat{\includegraphics[width = 1.5in]{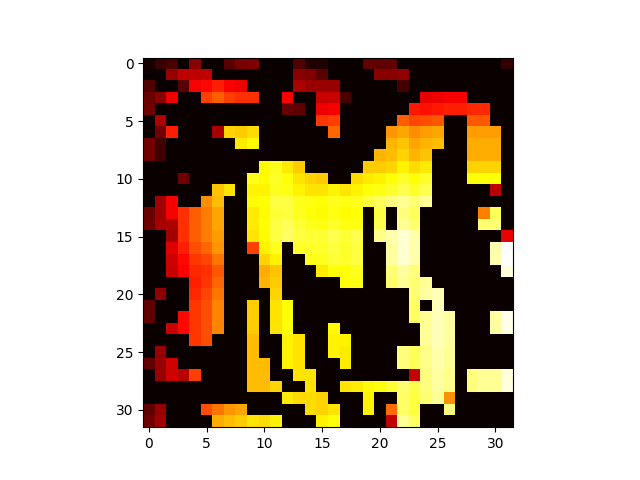}} \\
			\subfloat{\includegraphics[width = 1.5in]{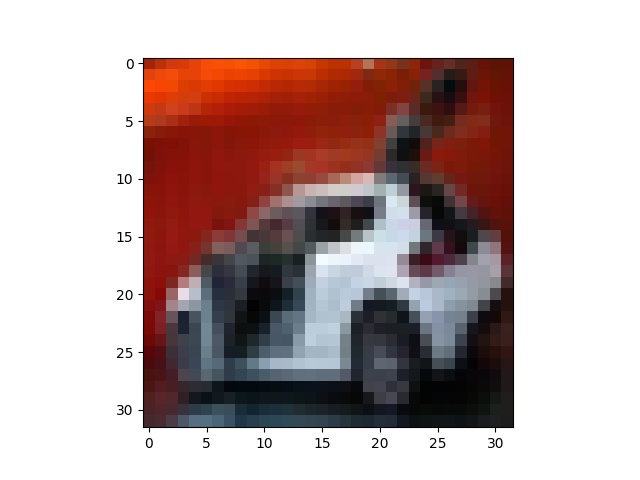}} &
			\subfloat{\includegraphics[width = 1.5in]{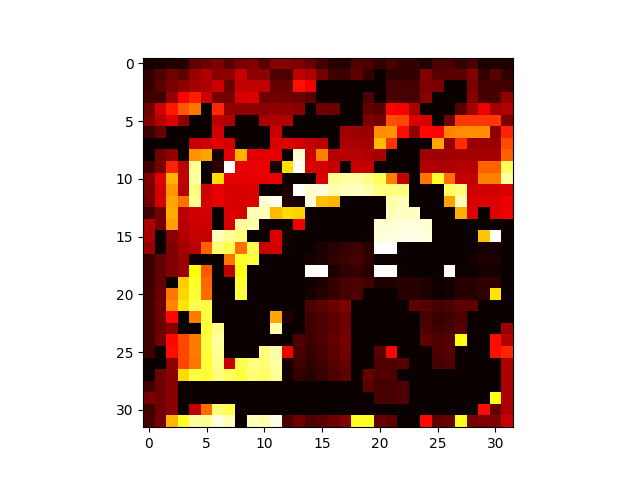}} &
			\subfloat{\includegraphics[width = 1.5in]{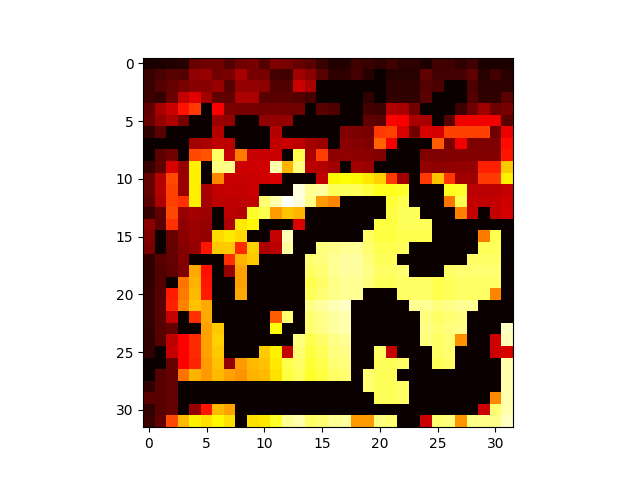}} \\
			\subfloat{\includegraphics[width = 1.5in]{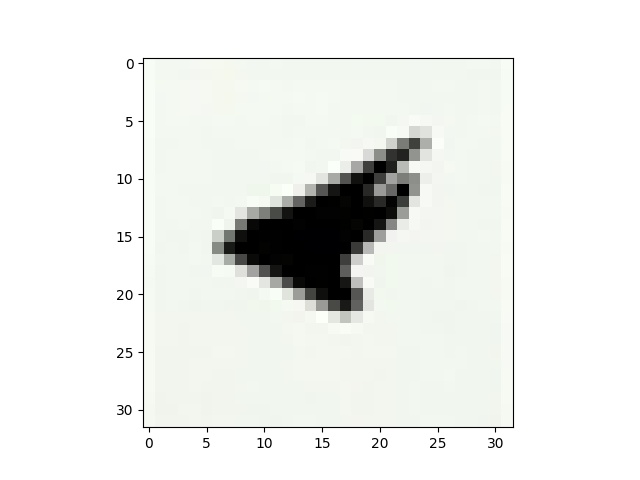}} &
			\subfloat{\includegraphics[width = 1.5in]{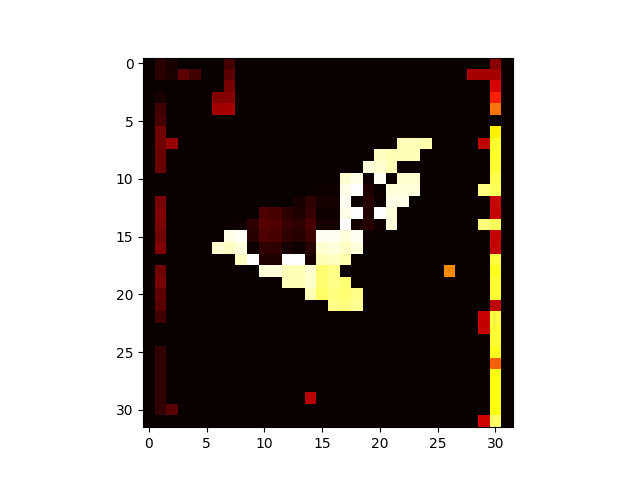}} &
			\subfloat{\includegraphics[width = 1.5in]{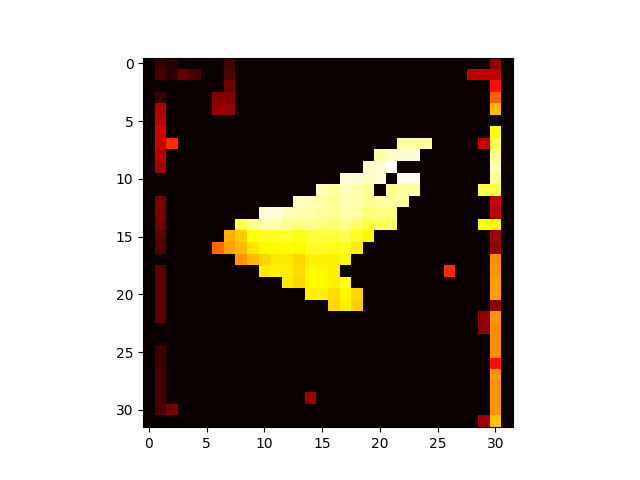}} \\
			\subfloat{\includegraphics[width = 1.5in]{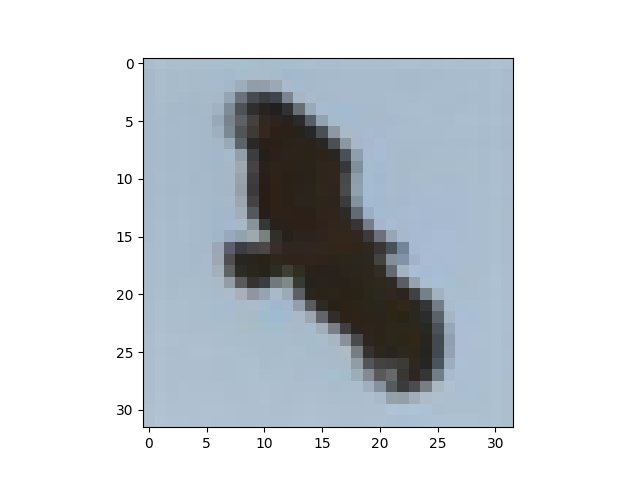}} &
			\subfloat{\includegraphics[width = 1.5in]{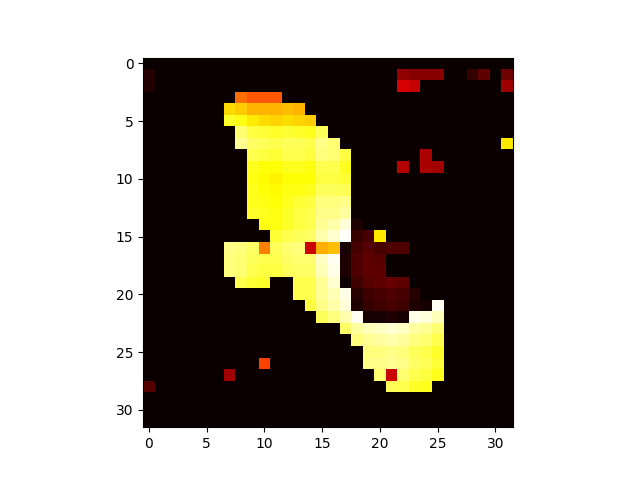}} &
			\subfloat{\includegraphics[width = 1.5in]{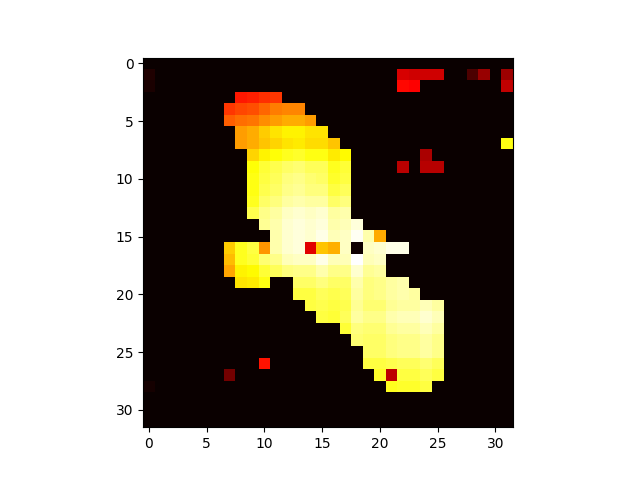}} \\
			\subfloat{\includegraphics[width = 1.5in]{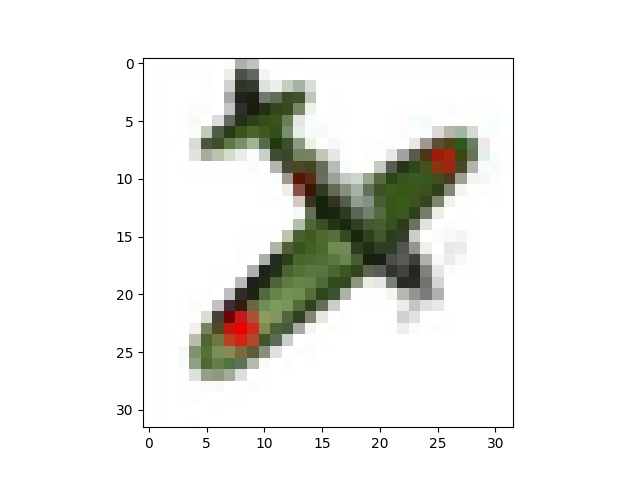}} &
			\subfloat{\includegraphics[width = 1.5in]{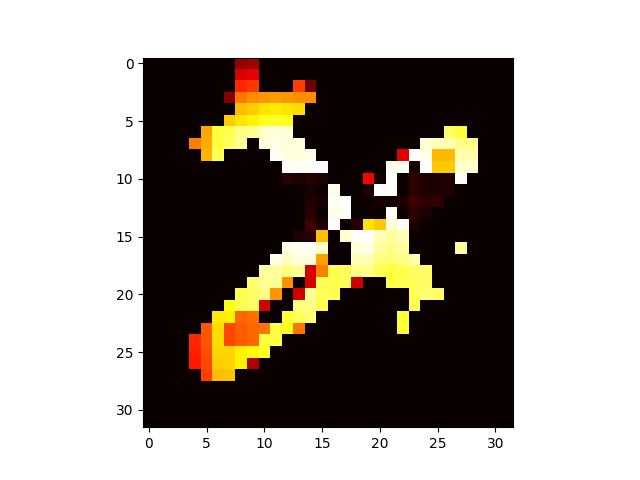}} &
			\subfloat{\includegraphics[width = 1.5in]{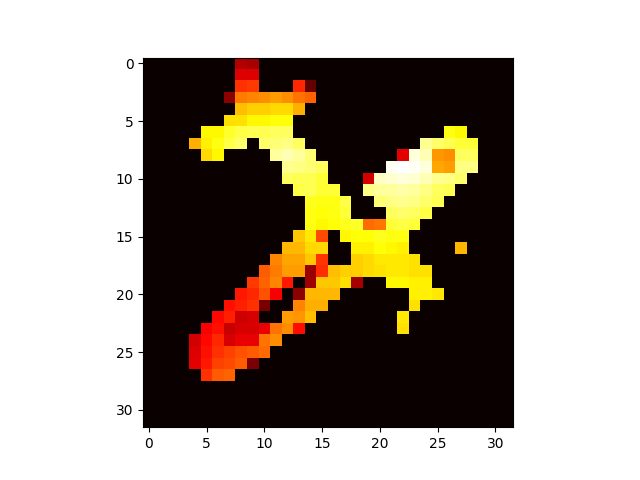}} \\
			\subfloat{\includegraphics[width = 1.5in]{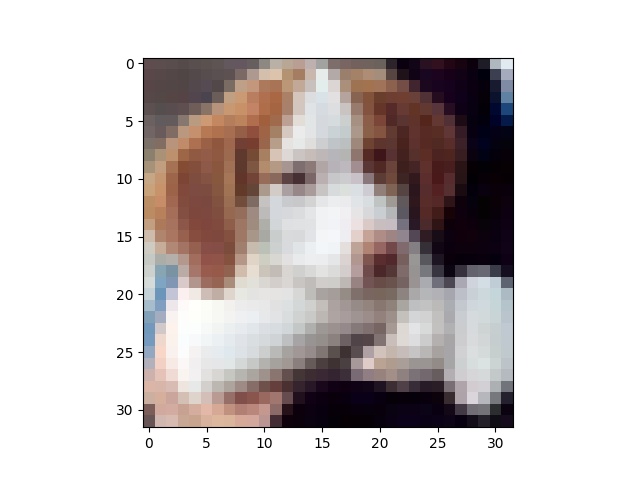}} &
			\subfloat{\includegraphics[width = 1.5in]{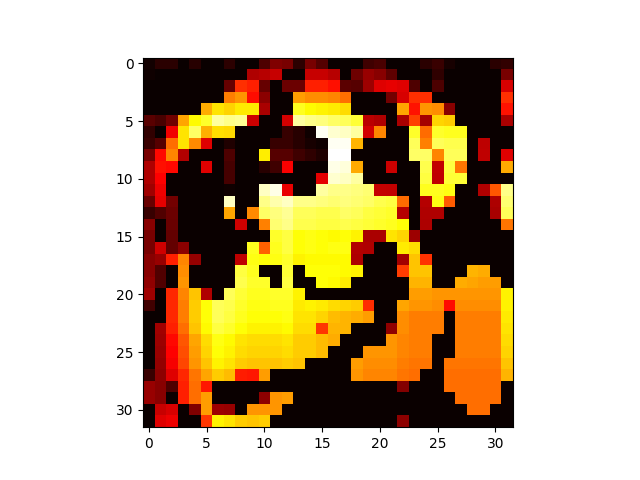}} &
			\subfloat{\includegraphics[width = 1.5in]{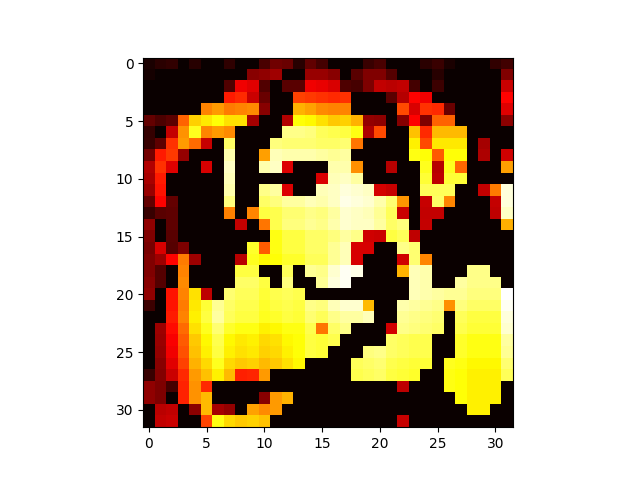}} \\
		\end{tabular}
		\caption{CIFAR Heatmaps}
	\end{figure}


\end{document}